\documentclass{article} 
\usepackage{iclr2027_conference,times}

\usepackage{amsmath,amsfonts,bm}

\def\eqref#1{equation~\ref{#1}}

\def\1{\bm{1}}

\DeclareMathAlphabet{\mathsfit}{\encodingdefault}{\sfdefault}{m}{sl}
\SetMathAlphabet{\mathsfit}{bold}{\encodingdefault}{\sfdefault}{bx}{n}

\usepackage{hyperref}
\usepackage{url}
\usepackage{graphicx}
\usepackage{wrapfig}
\usepackage{needspace}
\usepackage{capt-of}
\usepackage{booktabs}
\usepackage{pifont}
\usepackage{color}
\usepackage{enumitem}

\definecolor{rarTop}{HTML}{C01F6E}
\definecolor{rarMid}{HTML}{3B4DB7}
\definecolor{rarBottom}{HTML}{2FA34A}
\definecolor{citecolor}{RGB}{66,168,235}
\definecolor{linkcolor}{RGB}{255,0,0}

\newcommand{\RARlogo}{\textbf{\textcolor{rarTop}{R}\textcolor{rarMid}{A}\textcolor{rarBottom}{R}}}

\newcommand{\ie}{\textit{i.e.}}

\makeatletter
\renewcommand{\footnoterule}{\kern-3\p@ \hrule\@width\columnwidth \kern2.6\p@}
\makeatother

\hypersetup{colorlinks=true,citecolor=citecolor,linkcolor=linkcolor}

\title{\RARlogo: Rate-Adaptive Rollout for \\
Time-Consistent Interactive World Models}

\author{%
\textbf{Yaoyang Liu}$^{1}$\thanks{Equal contribution. \hspace{1.5em} $^{\dagger}$Corresponding author.} \quad
\textbf{Dongyu Yan}$^{2}$\footnotemark[1] \quad
\textbf{Yihua Du}$^{2}$ \quad
\textbf{Shengju Qian}$^{3}$ \quad
\textbf{Zeke Xie}$^{2}$ \quad
\textbf{Xin Wang}$^{3}$ \\
\textbf{Ying-Cong Chen}$^{2}$ \quad
\textbf{Long Chen}$^{1\dagger}$ \\[0.5em]
$^{1}$HKUST \quad
$^{2}$HKUST(GZ) \quad
$^{3}$LIGHTSPEED
}

\iclrfinalcopy

\begin{document}

\maketitle
\lhead{}
\renewcommand{\headrulewidth}{0pt}


\begin{abstract}
Video-based interactive world models aim to enable real-time, user-controllable environments.
Yet most existing systems are trained and rolled out at a fixed temporal rate, limiting both runtime-rate adaptation and physical-horizon coverage---a limitation we term \textbf{temporal-rate coupling}.
Specifically, when runtime generation throughput differs from the learned temporal rate, the generated world evolves unintentionally more slowly or quickly.
Moreover, using a single temporal rate imposes an inherent trade-off between temporal granularity and physical horizon: higher rates capture finer motion but cover a shorter duration under a fixed context window.
To address both limitations, we introduce \textbf{\textcolor{rarTop}{R}ate-\textcolor{rarMid}{A}daptive \textcolor{rarBottom}{R}ollout} (\RARlogo), a framework that enables interactive world models to roll out at controllable temporal rates.
At its core, \emph{Stride-RoPE} aligns latent frames sampled at different rates on a shared physical-time coordinate, enabling consistent multi-rate rollout, unseen-rate generalization, and online rate switching without discarding history.
Meanwhile, RAR uses multi-stride video sampling to enable joint learning of fine-grained dynamics and long-horizon evolution.
Across action- and camera-controlled environments, RAR achieves temporally consistent multi-rate rollout and online rate switching while preserving high-rate generation quality and extending physical horizons under matched training-update budgets. RAR also transfers to pretrained world models while preserving perceptual quality.
\end{abstract}

\section{Introduction}

Interactive world models are extending video generation from offline content synthesis toward real-time, user-controllable environments~\citep{bruce2024genie,he2025matrix,solaris2026,hu2026multiplayerinteractiveworldmodels}.
Unlike conventional video generation, these models must continuously receive user actions or camera trajectories and promptly synthesize subsequent visual observations consistent with the control inputs. The generated content then becomes part of the history that conditions future evolution.
Recent systems have demonstrated high-quality, long-horizon, and even real-time interactive generation in game and open-world settings~\citep{valevski2025diffusion,wang2026matrix,chen2026reworld,lingbot-world-v2}, highlighting their potential as neural game engines and general-purpose environment simulators.
As these systems move toward practical deployment, they must advance the generated world at the intended pace under varying runtime conditions, while also capturing both fine-grained dynamics and long-horizon evolution within a limited temporal context.

\begin{figure*}[t!]
    \centering
    \includegraphics[width=\textwidth]{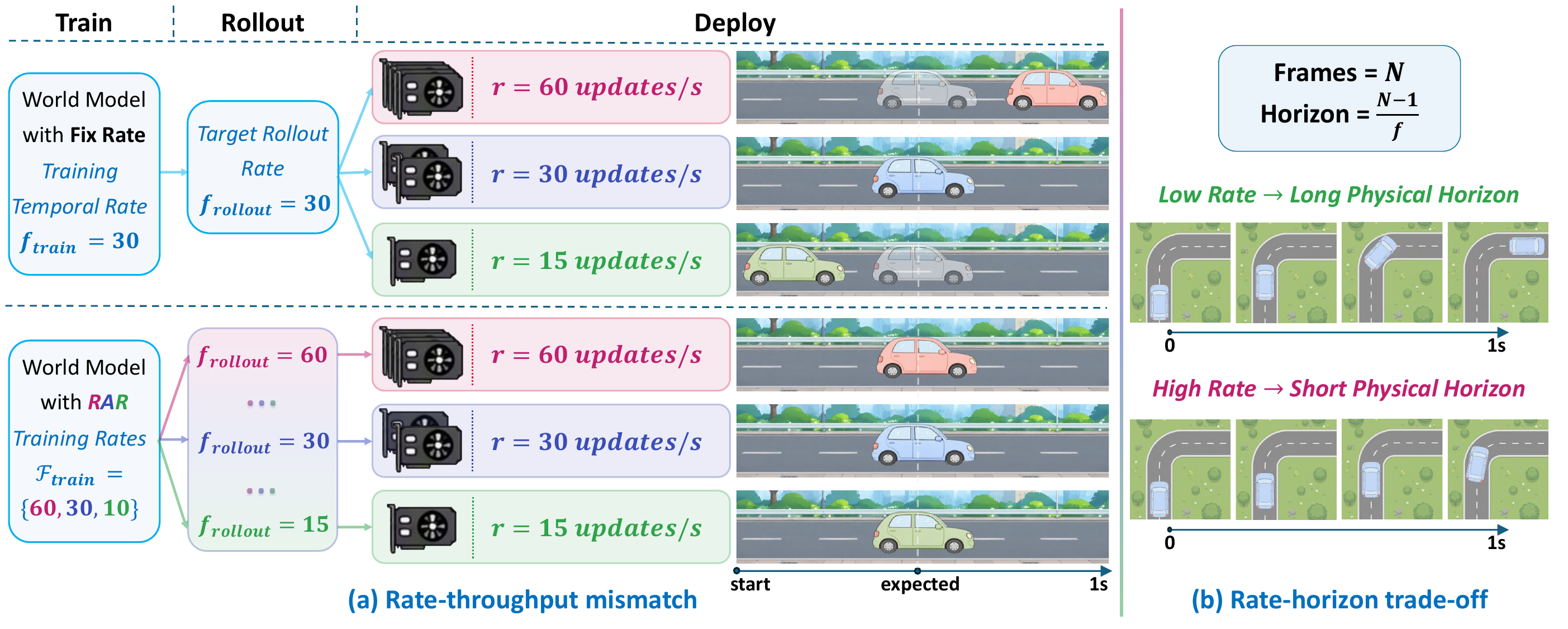}
    \caption{\textbf{(a)}: \emph{Rate--throughput mismatch}. $f_{\mathrm{train}}$, $f_{\mathrm{rollout}}$, and $r$ denote the training temporal rate, target rollout rate, and runtime generation throughput. After 1s of wall-clock time, colored cars mark generated positions, while \textcolor{gray}{gray} cars mark the \textit{expected} positions; their offsets indicate temporal mismatch. \textbf{(b)}: \emph{Rate--horizon trade-off}. Under the same fixed budget of $N$ frames, a higher rate provides finer temporal resolution but covers a shorter physical horizon, whereas a lower rate uses larger temporal steps to cover a longer horizon.}
    \label{fig:rar_problem}
\end{figure*}
Nevertheless, most existing interactive world models are trained and rolled out at a single fixed temporal rate~\citep{he2025matrix,wang2026matrix,solaris2026}.
We use the temporal rate $f$ to denote the number of temporal steps represented per second of physical time\footnote{In our video-based setting, temporal steps correspond to video or latent frames, so $f$ is instantiated as frames per second (FPS), and each temporal step represents a physical-time interval of $\Delta t=1/f$ seconds.}.
In a conventional fixed-rate model, the training rate and target rollout rate are tied together, \ie, $f_{\mathrm{train}}=f_{\mathrm{rollout}}=f$.
We argue that this fixed-rate design gives rise to two closely related limitations in world modeling:
\begin{itemize}[left=0.5em, labelsep=0.5em]
    \item \textbf{Rate--throughput mismatch.}
    With a runtime throughput of $r$ generation updates per wall-clock second, the modeled world advances at a relative speed of $\rho=r/f_{\mathrm{rollout}}$.
    For users, holding a movement command for the same wall-clock duration can therefore produce too little or too much displacement, making navigation and action timing inconsistent across runtime conditions.
    As shown in Fig.~\ref{fig:rar_problem}(a), the generated world lags behind or runs ahead of its expected state when $r<f_{\mathrm{rollout}}$ or $r>f_{\mathrm{rollout}}$, respectively.
    Although variable frame rates have been explored in offline video generation~\citep{ho2022video,blattmann2023stable,hacohen2026ltx}, these methods do not address adapting the physical-time progression of continuous interactive rollouts to runtime generation throughput.
    Moreover, adjusting playback speed only changes display timing, not the physical-time interval represented by each generated transition; addressing this mismatch therefore requires controlling the temporal step size within the world model itself.
\item \textbf{Rate--horizon trade-off.}
    With a fixed context budget of $N$ temporal frames, the modeled physical horizon is $T=(N-1)/f$.
    As shown in Fig.~\ref{fig:rar_problem}(b), a higher $f$ captures finer motion and control responses but covers a shorter duration, whereas a lower $f$ extends the horizon with coarser temporal steps.
    This constrains the model's ability to learn both rapid control responses and longer-term environment evolution.
    Recovering a long horizon at a high temporal rate requires longer training sequences and substantially higher training cost per optimization update.
\end{itemize}

Although the two problems arise in different aspects of interactive world modeling, both stem from assigning a fixed physical-time increment to every model update.
We refer to these two consequences collectively as \textbf{temporal-rate coupling}.
Breaking this coupling requires learning controlled dynamics across temporal rates within a shared physical-time coordinate.

To this end, we introduce \textbf{\textcolor{rarTop}{R}ate-\textcolor{rarMid}{A}daptive \textcolor{rarBottom}{R}ollout} (\RARlogo), which enables an interactive world model to be trained and rolled out across multiple temporal rates.
RAR parameterizes the temporal rate by a stride $s$ consistently during training and inference.
To learn across different strides, the model must preserve consistent temporal relations between frames separated by the same physical-time interval. We therefore introduce \emph{Stride-RoPE}, which modifies the original RoPE ~\citep{su2024roformer}, and maps latent frames sampled with different strides onto a shared physical-time coordinate for both visual observations and control conditioning.
During training, RAR synchronously samples video frames and control signals at multiple temporal strides, allowing the model to learn environment rollouts at varying temporal rates $\mathcal{F}_{\mathrm{train}}$ by adjusting the stride $s\in\mathcal{S}_{\mathrm{train}}$.
This training paradigm also allows fine-rate samples to capture rapid local dynamics and control responses while coarse-rate samples cover longer physical horizons under the same sequence budget.
At inference, the stride $s$ directly controls the target rollout rate  $f_{\mathrm{rollout}}$, enabling the model to adapt its physical-time progression to runtime generation throughput.
The stride can vary continuously within the learned range, including unseen intermediate rates, and can change online without discarding rollout history.

Across action- and camera-controlled interactive environments, RAR enables temporally consistent and controllable multi-rate rollout, including generalization to unseen rates and online rate switching, while extending physical horizons without sacrificing high-rate generation quality under matched training-update budgets.
RAR also transfers effectively to pretrained world models through finetuning, demonstrating its portability.
Together, these results demonstrate that RAR effectively mitigates temporal-rate coupling by enabling controllable rollout rates to adapt world progression to runtime generation throughput and by extending physical horizons under a fixed sequence budget. 
\section{Related Work}

\noindent\textbf{Interactive World Models.}
World models learn predictive environment dynamics to anticipate future observations and support decision making~\citep{ha2018world}.
Recent video-based world models have progressed from action-conditioned future generation~\citep{hu2023gaia} toward continuously controllable environments that support interactive rollout through user actions or camera trajectories~\citep{bruce2024genie,sun2026worldplaylongtermgeometricconsistency,hyworld2025,chen2026reworld,lingbot-world-v2,huang2026solarwm}.
Recent game- and open-world systems further scale this paradigm to more persistent closed-loop interaction, with stronger action--state coupling and improved long-horizon rollout consistency~\citep{valevski2025diffusion,po2026multigenleveldesigneditablemultiplayer,he2025matrix,wang2026matrix,solaris2026,hu2026multiplayerinteractiveworldmodels,team2026dreamx}.
Despite these advances, existing systems generally associate each rollout update with a fixed temporal interval. 
Our work instead considers temporal rate itself as a controllable dimension of interactive rollout, rather than a fixed property.

\noindent\textbf{Variable Frame-Rate Video Generation.}
Video generation models have explored varying temporal density through temporal resampling, super-resolution, and explicit FPS conditioning~\citep{ho2022video,ho2022imagen,chen2023videocrafter1,xing2024dynamicrafter,blattmann2023stable}.
While FPS conditioning provides global rate control, it does not explicitly enforce shared temporal geometry across different rates.
More recent methods incorporate frame rate into temporal positional coordinates to obtain physical-time-aware representations~\citep{hacohen2024ltx,hacohen2026ltx}.
However, these approaches primarily target finite offline videos with globally specified frame rates, rather than streaming interactive rollouts with time-consistent history across online rate changes.

\noindent\textbf{Multi-Scale and Long-Horizon Video Modeling.}
Long-context methods extend the number of sequence positions available to a model for modeling longer-range dependencies through positional interpolation or RoPE rescaling~\citep{blocntkaware,peng2024yarn,ding2024longrope}.
In video modeling, temporal multi-scale and hierarchical approaches cover longer time spans by operating at multiple temporal resolutions or by generating sparse long-range structure before refining finer temporal details~\citep{villar2022mspred,du2023learning,yin2023nuwa}.
These approaches extend temporal coverage by enlarging context or through hierarchical generation, whereas our setting varies the physical duration represented by a fixed-length rollout context.

\section{Method}
\begin{figure}[t]
    \centering
    \includegraphics[width=\linewidth]{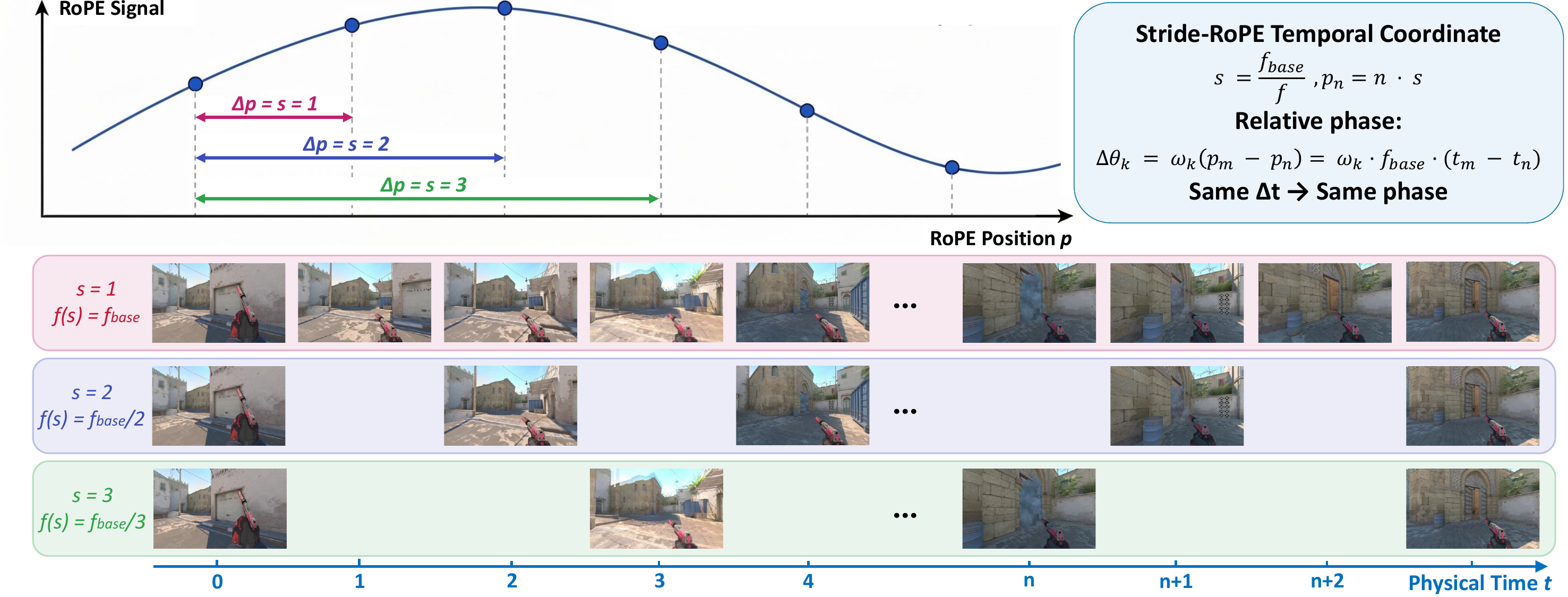}
    \caption{\emph{Stride-RoPE}. Video sampled with temporal stride $s$ uses RoPE positions $p_n=ns$, ensuring that equal physical-time intervals induce equal relative phases across sampling rates.}
    \label{fig:rar_method}
\end{figure}

We introduce \textbf{Rate-Adaptive Rollout (RAR)}, a framework for multi-rate training and rollout of interactive world models. RAR is built around a unified temporal stride that parameterizes physical-time progression across training and inference. This stride defines rate-consistent temporal coordinates through \emph{Stride-RoPE} (Sec.~\ref{sec:stride_rope}) and governs multi-stride training that balances rate and horizon, rate-controlled rollout and online rate switching (Sec.~\ref{sec:rar}). 

\subsection{Problem Definition}
\label{sec:problem_definition}

We first formalize the problem. 
Let $\mathbf{x}_n=\mathbf{x}(t_n)$ denote the world state at physical time $t_n$, and let $\mathbf{a}_n$ denote the control applied during the transition from $\mathbf{x}_n$ to $\mathbf{x}_{n+1}$.
Given the history of world states and controls, a causal interactive world model autoregressively predicts the next state as:
\begin{equation}
    \widehat{\mathbf{x}}_{n+1}
    \sim
    p_{\theta}\!\left(
        \cdot
        \mid
        \mathbf{x}_{0:n},
        \mathbf{a}_{0:n}
    \right).
    \label{eq:ar_rollout}
\end{equation}
Each transition advances the modeled world by a physical-time interval $\Delta t_n = t_{n+1} - t_n$.
Conventional models use a constant step size determined by the fixed training temporal rate, thereby tying each model update to a fixed amount of physical-time progression.
We instead require the world model to support controllable temporal intervals $\Delta t_n$, both across training samples and within an ongoing rollout.
Accordingly, we seek a variable-rate world model with two key properties.
First, different temporal discretizations should preserve consistent temporal semantics at matched physical times, while changing the step size during rollout should not alter the physical-time relationships of previously generated states.
Second, learning across multiple step sizes allows fine steps to capture fine-grained local dynamics and coarse steps to extend physical-time coverage under the same context budget.
These requirements call for a shared physical-time representation across rates and a unified mechanism for learning and controlling temporal intervals.

\subsection{Stride-RoPE}
\label{sec:stride_rope}

To support this physical-time consistency, we use a temporal positional representation in which equal physical-time intervals induce equal relative phases across sampling rates. 
We first examine the temporal geometry of RoPE, which encodes positions by rotating query and key subspaces. With frame-index-based temporal coordinates, the interaction between a query at index $n$ and a key at index $m$ in the $k$-th rotary subspace satisfies:
\begin{equation}
\left\langle
R(n\omega_k)\mathbf{q}_{n,k},
R(m\omega_k)\mathbf{k}_{m,k}
\right\rangle
=
\left\langle
\mathbf{q}_{n,k},
R\!\left((m-n)\omega_k\right)\mathbf{k}_{m,k}
\right\rangle,
\label{eq:rope_relative}
\end{equation}
where $\omega_k$ is the angular frequency of that subspace. The relative rotary phase is therefore determined by the frame-index difference $m-n$. For a sequence sampled at a constant temporal rate $f$, this difference corresponds to the physical-time interval:
\begin{equation}
t_m-t_n=\frac{m-n}{f}.
\label{eq:index_time}
\end{equation}
The relative phase can thus be written as
$(m-n)\omega_k=f(t_m-t_n)\omega_k$.
Consequently, the phase assigned to a given physical-time interval depends on the sampling rate: frame-index-based RoPE does not provide a shared physical-time reference across rates.
Thus, the temporal positional encoding itself does not explicitly represent elapsed physical time.

To establish such a reference, we reparameterize the temporal RoPE coordinate in units of physical time. Given a base temporal rate $f_{\mathrm{base}}$, we define the stride and the corresponding temporal coordinate for a sequence sampled at a constant rate $f$ as:
\begin{equation}
s(f)=\frac{f_{\mathrm{base}}}{f},
\qquad
p_n=n s(f).
\label{eq:stride_position}
\end{equation}
Instead of applying RoPE at the frame index $n$, \emph{Stride-RoPE} applies it at the rescaled coordinate $p_n$. Taking $t_0=0$, we have $t_n=n/f$ and hence,
\[
p_n=f_{\mathrm{base}}t_n.
\]
Thus, $p_n$ measures physical time in units of the base-frame interval $1/f_{\mathrm{base}}$. The relative rotary phase becomes:
\begin{equation}
(p_m-p_n)\omega_k
=(m-n)\frac{f_{\mathrm{base}}}{f}\omega_k
=f_{\mathrm{base}}(t_m-t_n)\omega_k.
\label{eq:stride_relative_phase}
\end{equation}
For a fixed base rate, this phase depends on the physical-time interval rather than the sampling rate $f$. 
Equal physical-time intervals therefore induce equal relative rotary phases across rates, establishing a shared temporal positional geometry for multi-rate learning.

When $f=f_{\mathrm{base}}$, we have $s=1$ and $p_n=n$, so Stride-RoPE exactly recovers standard RoPE at the base rate and preserves its original positional geometry. The modification affects only the temporal coordinates of 3D RoPE: the rotary frequencies and spatial coordinates remain unchanged, and no additional learnable parameters are introduced. Wherever temporal RoPE is applied to visual observations and control conditioning, we use the same coordinate mapping to maintain a common physical-time reference for both.

Although training trajectories are sampled using a discrete set of strides, the coordinate $p_n=ns$ is continuous in $s$ and well-defined for any real-valued $s>0$. It therefore provides temporal coordinates for intermediate strides not explicitly observed during training. 
We next describe how RAR builds on Stride-RoPE to address temporal-rate coupling.

\subsection{Rate-Adaptive Rollout Framework}
\label{sec:rar}

\begin{wrapfigure}{r}{0.52\linewidth}
    \centering
    \vspace{-1\baselineskip}
    \includegraphics[width=0.52\textwidth]{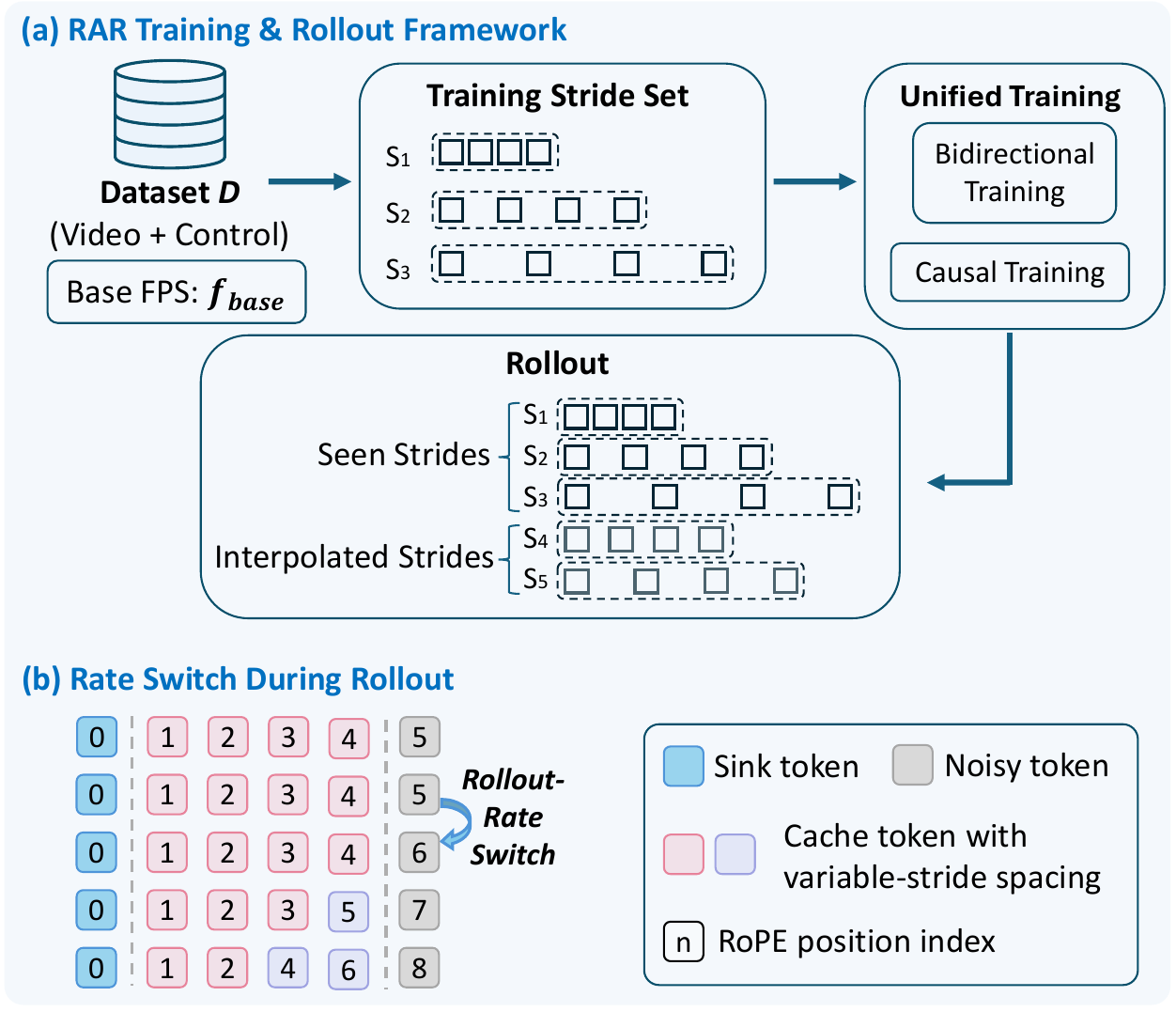}
    \vspace{-1.6\baselineskip}
    \caption{RAR pipeline. \textbf{(a)} Multi-stride training and rate-controlled rollout. \textbf{(b)} Rollout-rate switching with stride-aware temporal spacing.}
    \label{fig:rar_pipeline}
    \vspace{-2.1\baselineskip}
\end{wrapfigure}

RAR uses the stride variable to unify multi-rate training and rollout, with online rate switching achieved by varying the stride within an ongoing rollout. Fig.~\ref{fig:rar_pipeline} summarizes our framework.

\paragraph{Multi-rate training.}
We choose a set of training strides $\mathcal{S}_{\mathrm{train}}$ with respect to the fixed base temporal rate $f_{\mathrm{base}}$, yielding the rate set
$\mathcal{F}_{\mathrm{train}}=\{f_{\mathrm{base}}/s\mid s\in\mathcal{S}_{\mathrm{train}}\}$.
For a sampled stride $s\in\mathcal{S}_{\mathrm{train}}$, we synchronously sample video and its associated control signals at the following physical times:
\begin{equation}
    t_n^{(s)}=\frac{ns}{f_{\mathrm{base}}},
    \qquad
    \mathbf{x}_n^{(s)}=\mathbf{x}\!\left(t_n^{(s)}\right),
    \qquad
    \mathbf{a}_n^{(s)}=\mathbf{a}\!\left(t_n^{(s)}\right).
    \label{eq:multirate_sampling}
\end{equation}
Here, $n=0,\ldots,N-1$ indexes video frames and $n=0,\ldots,N-2$ indexes transition controls.
The notation $\mathbf{a}(t_n^{(s)})$ denotes the control representation for the interval $[t_n^{(s)},t_{n+1}^{(s)})$, consistent with Eq.~\ref{eq:ar_rollout}, rather than point sampling all control channels.
From the base-rate control stream, held-key states are sampled synchronously with the stride, events are triggered if pressed within the interval, mouse increments are summed over the interval, and camera trajectories use reference-relative poses at the sampling times.
The same rules apply at training and interpolated strides, with non-integer sample positions mapped to the nearest base-rate frame.

Every training sequence contains the same $N$ video frames, with a uniform temporal interval $\Delta t(s)=s/f_{\mathrm{base}}$, while its physical-time coverage is
\begin{equation}
    T(s)=(N-1)\Delta t(s)=\frac{(N-1)s}{f_{\mathrm{base}}}.
    \label{eq:temporal_coverage}
\end{equation}
Fine-rate samples capture fine-grained local motion and control responses, whereas coarse-rate samples cover longer physical horizons under the same sequence budget. Joint training across these rates therefore enables learning at multiple temporal scales without increasing the training sequence length.
This stride-controlled sampling strategy can be readily applied to different training stages, including bidirectional, causal training, and distillation.

\paragraph{Rate-controlled rollout.}
At inference, RAR allows the temporal interval to be selected at each autoregressive step.
Let $s_n$ denote the stride for the transition from $\mathbf{x}_n$ to $\mathbf{x}_{n+1}$, corresponding to the interval $\Delta t_n=s_n/f_{\mathrm{base}}$.
For a constant stride $s_n=s_{\mathrm{rollout}}$ and runtime
generation throughput $r$, the target rollout rate and relative
physical-time progression are
\begin{align}
    f_{\mathrm{rollout}}
    &=\frac{f_{\mathrm{base}}}{s_{\mathrm{rollout}}},
    \qquad
    \rho=\frac{r}{f_{\mathrm{rollout}}}
    =\frac{r\,s_{\mathrm{rollout}}}{f_{\mathrm{base}}}.
    \label{eq:runtime_alignment}
\end{align}
Together with multi-rate training, the continuous stride parameterization of Stride-RoPE supports rollout at unseen intermediate rates within the training range.
When $r$ falls within this range, setting $s_{\mathrm{rollout}}=f_{\mathrm{base}}/r$ matches the target  rollout rate to runtime throughput, yielding $\rho=1$.

More generally, the stride can vary from one transition to the next within the supported range, allowing the rollout rate to change online.
We compute the temporal RoPE coordinate $p_n$ of frame $n$ by accumulating the selected strides:
\begin{equation}
    \begin{gathered}
        p_0=0,
        \qquad
        p_{n+1}=p_n+s_n,
        \\
        p_{\ell}-p_i
        =\sum_{j=i}^{\ell-1}s_j
        =f_{\mathrm{base}}(t_{\ell}-t_i),
        \qquad i<\ell.
    \end{gathered}
    \label{eq:cumulative_positions}
\end{equation}
For non-sink frame pairs, the relative rotary phase remains $(p_{\ell}-p_i)\omega_k=f_{\mathrm{base}}(t_{\ell}-t_i)\omega_k$, extending the physical-time alignment in Eq.~\ref{eq:stride_relative_phase} to rollouts with time-varying strides.

\begin{figure}[t]
    \centering
    \includegraphics[width=\linewidth]{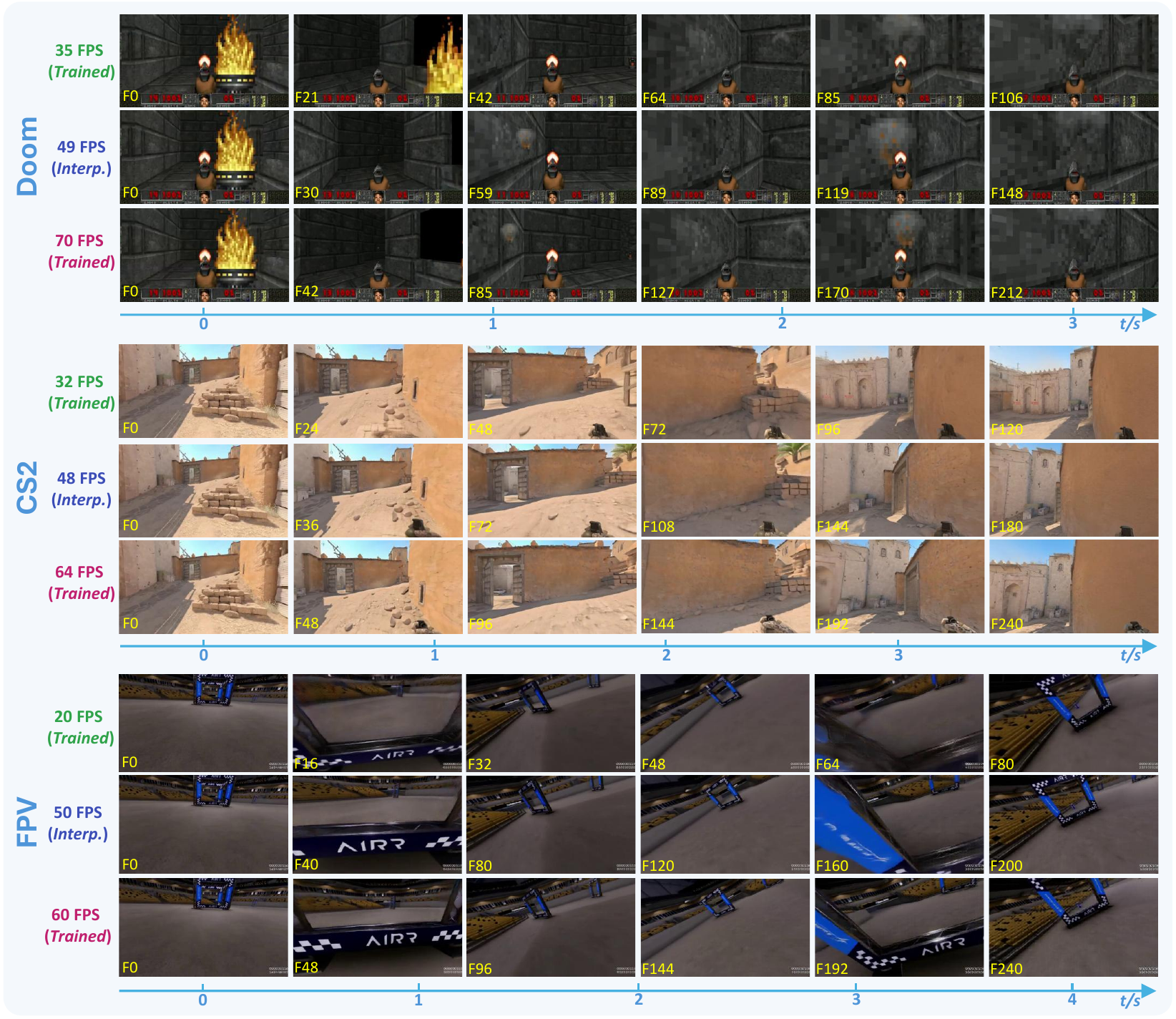}
    \caption{Qualitative multi-rate generation. Within each environment, rows share the same initial condition and underlying physical-time control trajectory, while columns are aligned by physical time. Frame indices are marked at the lower left of each frame. At matched physical times, RAR produces similar scene states across rollout rates.}
    \label{fig:qualitative_multirate}
    \vspace{-0.1\baselineskip}
\end{figure}

We implement causal rollout with a KV cache following the \emph{Unified Relative RoPE Recipe}~\citep{bian2026echoinfinity}.
The cache stores pre-RoPE keys together with their corresponding values.
Sink frames retain fixed RoPE positions following the adopted cache recipe.
For the current frame $\ell$ and a cached frame $i$, temporal RoPE in the $k$-th rotary subspace is applied as:
\begin{equation}
    \widetilde{\mathbf{q}}_{\ell,k}
    =R(p_{\ell}\omega_k)\mathbf{q}_{\ell,k},
    \qquad
    \widetilde{\mathbf{k}}_{i,k}
    =R(p_i\omega_k)\mathbf{k}^{\mathrm{raw}}_{i,k}.
    \label{eq:relative_cache_rope}
\end{equation}
At each attention step, the coordinates of active frames are determined from the accumulated stride history.
These coordinates may be translated by a common offset within the active window without changing their relative rotary phases.
Applying the resulting rotations directly to the cached raw keys allows the rollout rate to change without clearing the cache or recomputing historical features.

Although training uses a uniform stride within each sequence, online switching combines stride increments within the training range into mixed histories.
For non-sink frames, the relative rotary phase depends only on physical-time differences, while $p_{n+1}-p_n=s_n$ specifies the next time advance.
Changing this increment does not alter cached past states or the physical-time differences between existing non-sink frames.
Our switching results (Appd.~\ref{app:rate_switch}) demonstrate mixed-stride rollout without dedicated mixed-stride training.

\section{Experiments}

\begin{table*}[t]
    \centering
    \caption{Multi-rate generation under a fixed rollout budget.  Drift is the ratio between last-quarter and first-quarter LPIPS. DOVER is the no-reference overall perceptual quality score in $[0,1]$; higher is better. $\mathrm{SR}$ measures rate fidelity and has an ideal value of 1.}
    \label{tab:main_results}
    \small
    \setlength{\tabcolsep}{5pt}
    \renewcommand{\arraystretch}{0.9}
    \begin{tabular*}{\textwidth}{@{\extracolsep{\fill}}lccccccc@{}}
        \toprule
        Environment & FPS & Type & Horizon (s) & LPIPS $\downarrow$ & Drift & DOVER $\uparrow$ & $\mathrm{SR}\!\rightarrow\!1$ \\
        \midrule
        Doom & 14 & Train   & 15.1 & 0.4279 & 1.17 & 0.3205 & 0.923 \\
             & 28 & \textbf{Interp.} &  7.6 & 0.4179 & 1.22 & 0.3269 & 1.003 \\
             & 35 & Train   &  6.1 & 0.3991 & 1.27 & 0.3153 & 0.999 \\
             & 49 & \textbf{Interp.} &  4.3 & 0.4003 & 1.26 & 0.3156 & 0.986 \\
             & 70 & Train   &  3.0 & 0.3852 & 1.34 & 0.3049 & 1.082 \\
        \midrule
        CS2  & 16 & Train   & 15.0 & 0.6489 & 1.52 & 0.3928 & 0.983 \\
             & 24 & \textbf{Interp.} & 10.0 & 0.6200 & 1.60 & 0.4187 & 1.009 \\
             & 32 & Train   &  7.5 & 0.5980 & 1.61 & 0.4402 & 0.970 \\
             & 48 & \textbf{Interp.} &  5.0 & 0.5944 & 1.67 & 0.4369 & 1.024 \\
             & 64 & Train   &  3.8 & 0.5823 & 1.69 & 0.4323 & 1.097 \\
        \midrule
        FPV  & 20 & Train   & 12.0 & 0.5247 & 1.35 & 0.2827 & 1.031 \\
             & 30 & Train   &  8.0 & 0.5161 & 1.41 & 0.2766 & 1.040 \\
             & 40 & \textbf{Interp.} & 6.0 & 0.5011 & 1.42 & 0.2733 & 0.960 \\
             & 50 & \textbf{Interp.} &  4.8 & 0.5035 & 1.45 & 0.2843 & 1.007 \\
             & 60 & Train   &  4.0 & 0.4944 & 1.49 & 0.2956 & 0.917 \\
        \bottomrule
    \end{tabular*}
\end{table*}

In this section, we first introduce the experimental setup and then evaluate RAR's rate control, training benefits, and portability.
To address the rate--throughput mismatch, we examine two levels of rate control: \textbf{Q1:} Can RAR generate at specified rates, including unseen rates, while preserving visual quality and temporal consistency?
\textbf{Q2:} Can it further switch rates online without interrupting the rollout?
We then turn to the rate--horizon trade-off and ask \textbf{Q3:} Can multi-rate training alleviate this trade-off under matched training budgets?
Finally, we assess portability: \textbf{Q4:} Can RAR enable multi-rate rollout in pretrained world models without training from scratch?

\subsection{Experimental Setup}
\noindent\textbf{Environments.} We evaluate RAR in three interactive environments: Doom~\citep{idsoftware1993doom}, Counter-Strike 2 (CS2)~\citep{valve2023cs2}, and FPV~\citep{pfeiffer2021human}.
Doom and CS2 are first-person interactive games controlled through discrete keyboard-and-mouse actions, whereas FPV, a first-person drone racing simulator, involves continuous camera-trajectory control.

\noindent\textbf{Training and evaluation.}
We follow Solaris's three-stage training pipeline~\citep{solaris2026}, initializing Doom and CS2 models from Matrix-Game 2.0~\citep{he2025matrix} and FPV from SkyReels~\citep{chen2025skyreelsv2infinitelengthfilmgenerative}.
For each environment, we train one model on its temporal-rate set $\mathcal{F}_{\mathrm{train}}$.
Training details are provided in Appd.~\ref{app:experimental_details}, with stage-wise training steps listed in Tab.~\ref{tab:appendix_main_settings}.
The multi-rate generation and dynamic switching evaluations (Secs.~4.2.1 and~4.2.2) use the final checkpoints after all three training stages. The controlled rate--horizon comparisons in Tab.~\ref{tab:training_efficiency} (Sec.~4.3) use bidirectional checkpoints, complemented by final causal-model validation in Appd.~\ref{app:rate_control_comparison} and Appd.~\ref{app:causal_long_rollout}.
We evaluate each model at both training rates and unseen intermediate rates within the training range.
Cross-rate comparisons use matched initial conditions, random seeds, and the same underlying physical-time control trajectories.
Within each environment, we fix the number of generated frames across rate settings.

\noindent\textbf{Metrics.} We evaluate perceptual generation quality using the reference-based LPIPS~\citep{zhang2018unreasonable} and the no-reference DOVER overall score~\citep{wu2023exploring}.
To evaluate temporal semantics, we introduce the speed ratio $\mathrm{SR}$ as a rate-fidelity metric:
\begin{equation}
    \mathrm{SR}(f)=
    \frac{v_{\mathrm{pred}}(f)/\bar v_{\mathrm{pred}}}
    {v_{\mathrm{GT}}(f)/\bar v_{\mathrm{GT}}}.
    \label{eq:rate_fidelity}
\end{equation}
Here, $v(f)$ denotes motion speed estimated from RAFT~\citep{teed2020raft} optical flow using the physical-time interval between the compared frames, and $\bar v$ is its geometric mean across the rates within the training range.
A value of $\mathrm{SR}=1$ indicates that the generated and ground-truth videos exhibit the same cross-rate temporal scaling.
Values above or below 1 indicate over- or under-scaling of motion at that rate compared with the ground-truth cross-rate trend.
For SR evaluation, we filter out cases where optical flow estimation fails and remove spatial regions with excessively small or large motion magnitudes to reduce the influence of unreliable flow estimates.
The complete per-case definition, aggregation, and filtering procedure are provided in Appd.~\ref{app:rate_fidelity}.

\subsection{Rate-Controlled Rollout}

\subsubsection{Multi-Rate Generation (Q1)}

\noindent\textbf{Settings.} We compare rollouts at training and unseen intermediate rates, holding the target rate constant within each rollout.
Qualitative comparisons use frames at matched physical times, while quantitative evaluations fix the number of generated frames within each environment.

\noindent\textbf{Results.} As shown in Fig.~\ref{fig:qualitative_multirate}, RAR produces similar scene evolution and action progress at matched physical times across rollout rates.
This consistency also holds at unseen intermediate rates, supporting interpolation with the same model without retraining.

\begin{wrapfigure}{r}{0.52\linewidth}
    \centering
    \vspace{-0.5\baselineskip}
    \includegraphics[width=0.50\textwidth]{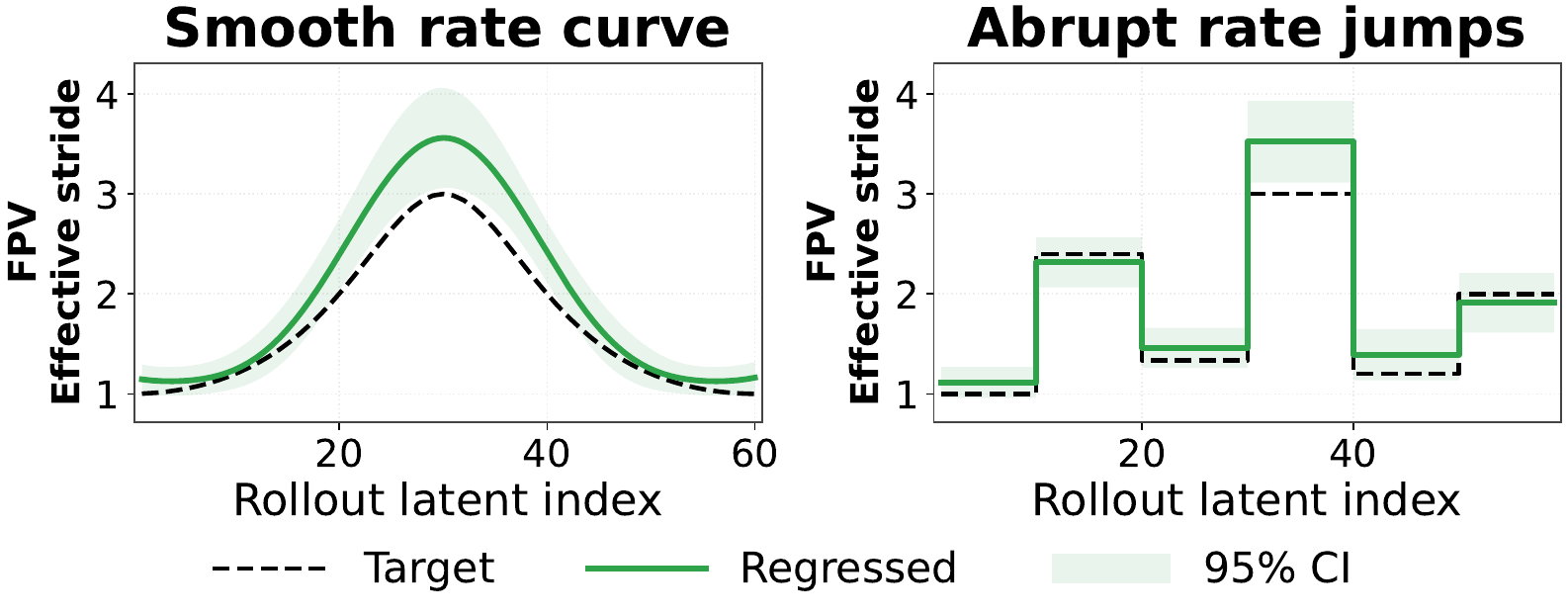}
    \caption{Dynamic rollout-rate switching results on FPV model. Dashed curves denote target strides; solid curves show the mean calibrated and regressed stride estimates over retained cases, with shading indicating 95\% confidence intervals.}
    \label{fig:rate_switch_main}
\end{wrapfigure}

Tab.~\ref{tab:main_results} reports quantitative results under a fixed rollout budget.
With the same number of generated frames, lowering the rollout rate from the highest to the lowest evaluated rate extends the physical horizon by factors of 5, 4, and 3 on Doom, CS2, and FPV, respectively.
These extensions are accompanied by modest LPIPS increases of 6.1\%--11.4\% relative to the respective highest-rate rollouts, while DOVER scores remain broadly comparable across rates within each environment.
Unseen intermediate rates also achieve LPIPS and DOVER scores comparable to neighboring training rates.
RAR also preserves consistent temporal scaling across rates.
The mean $|\mathrm{SR}-1|$ is 3.5\% on Doom, 3.5\% on CS2, and 4.0\% on FPV.
Across all six unseen intermediate rates, SR ranges from 0.960 to 1.024, with a maximum deviation of 4.0\% from the ideal value.
Their deviations are comparable to those of the training rates, indicating no systematic loss of rate fidelity at unseen rates.

Together, these results show that a single RAR model supports temporally consistent rollout across multiple rates and interpolates to unseen intermediate rates within the training range without rate-specific retraining.
Comparisons with alternative rate-control methods and results beyond the training range are provided in Appd.~\ref{app:rate_control_comparison} and Appd.~\ref{app:extrapolation}, respectively.

\subsubsection{Dynamic Rollout-Rate Switching (Q2)}

\textbf{Settings.} Beyond fixed-rate generation, we evaluate whether RAR can follow changing rollout rates within a continuous rollout without clearing the KV cache or restarting generation.
We use two stride schedules: a smoothly varying curve and an abrupt schedule comprising six constant-stride segments.
For each rollout, we rescale the optical-flow measurements to the target stride scale using a global factor before fitting the response.

\noindent\textbf{Results.} Fig.~\ref{fig:rate_switch_main} shows the mean fitted response on FPV and its 95\% confidence interval over the retained cases.
The reported Pearson correlations with the target schedules are 0.998 for the smooth schedule and 0.970 for the abrupt schedule.
These results support RAR's ability to track changes in the target stride during a continuous rollout.
Evaluation details and additional results on Doom and CS2 are provided in Appd.~\ref{app:rate_switch}.

\subsection{Rate--Horizon Trade-off (Q3)}

\textbf{Settings.} We conduct two comparisons at the bidirectional training stage on Doom.
(1) Matched updates: We fix the training window (213 frames), model architecture, and number of updates, comparing RAR with a single-rate model trained only at 70 FPS.
(2) Matched training time: We compare the 213-frame RAR model with fixed-rate models trained using 425- or 849-frame windows at 70 FPS under approximately matched wall-clock budgets.
For each matched-time comparison, both models are evaluated at 70 FPS using the same rollout length of 425 or 849 frames.

\textbf{Results.} (1) Under matched updates, RAR substantially improves generation quality at the lower rollout rates while retaining nearly identical performance at 70 FPS, as shown in Tab.~\ref{tab:training_efficiency} (top).
This indicates that exposing the model to longer physical horizons through coarse-rate samples does not come at the expense of its fine-rate generation capability.
(2) For the matched-time comparison, the 425- and 849-frame training configurations take $1.56\times$ and $5.25\times$ as long per update as the 213-frame configuration, respectively.
At the approximately matched checkpoints reported in Tab.~\ref{tab:training_efficiency} (bottom), RAR achieves lower LPIPS than the corresponding longer-window models.

These bidirectional results show that RAR alleviates the rate--horizon trade-off from two complementary perspectives: multi-rate training preserves fine-rate generation quality while exposing the model to longer physical-time contexts, and achieves long-horizon generation more effectively under an approximately matched training-time budget than directly lengthening high-rate training sequences.
Appd.~\ref{app:bidirectional_rate_horizon} provides the corresponding settings and additional results across bidirectional checkpoints.

\begin{table*}[t]
    \centering

    \begin{minipage}[t]{0.48\textwidth}
        \vspace{0pt}
        \centering
        \caption{Rate--horizon comparisons of bidirectional models on Doom: LPIPS under matched updates (top) and
        approximately matched wall-clock time (bottom). The baseline uses Stride-RoPE at inference.}
        \label{tab:training_efficiency}
    \end{minipage}\hfill
    \begin{minipage}[t]{0.48\textwidth}
        \vspace{0pt}
        \centering
        \caption{Multi-rate adaptation of SANA-WM.
        Baseline model directly applies Stride-RoPE at inference without
        additional finetuning.
        $\dagger$ marks the baseline model's native output rate.}
        \label{tab:sana_transfer}
    \end{minipage}
    \par\nointerlineskip
    \noindent\begin{minipage}[t]{0.48\textwidth}
        \vspace{0pt}
        \centering
        \small
        \renewcommand{\arraystretch}{1.1}
        \setlength{\tabcolsep}{3pt}
        \begin{tabular*}{\linewidth}
            {@{\extracolsep{\fill}}lccc@{}}
            \toprule
            Protocol & Evaluation & Baseline & RAR \\
            \midrule
            10k updates
                & 14 FPS & 0.4376 & \textbf{0.3834} \\
                & 35 FPS & 0.3863 & \textbf{0.3413} \\
                & 70 FPS & \textbf{0.3113} & 0.3135 \\
            \midrule
            Matched time
                & 425 frames & 0.3500 & \textbf{0.3383} \\
                & 849 frames & 0.3745 & \textbf{0.3552} \\
            \bottomrule
        \end{tabular*}
    \end{minipage}\hfill
    \begin{minipage}[t]{0.48\textwidth}
        \vspace{0pt}
        \centering
        \small
        \renewcommand{\arraystretch}{1.0}
        \setlength{\tabcolsep}{2pt}
        \begin{tabular*}{\linewidth}
            {@{\extracolsep{\fill}}lccccc@{}}
            \toprule
            & & \multicolumn{2}{c}{SR $\rightarrow 1$}
                & \multicolumn{2}{c}{LPIPS $\downarrow$} \\[-2pt]
            \cmidrule(lr){3-4}\cmidrule(l){5-6}
            FPS & Type & Base & RAR & Base & RAR \\[-2pt]
            \midrule
            10 & Train
               & 0.598 & \textbf{0.641}
               & 0.5702 & 0.4956 \\
            16\rlap{$^\dagger$} & \textbf{Interp.}
               & \textbf{1.203} & 1.238
               & 0.5684 & 0.5009 \\
            20 & Train
               & 0.983 & \textbf{0.989}
               & 0.5707 & 0.4979 \\
            24 & \textbf{Interp.}
               & \textbf{1.173} & \textbf{1.173}
               & 0.5673 & 0.5022 \\
            30 & Train
               & 1.234 & \textbf{1.127}
               & 0.5678 & 0.4999 \\
            \bottomrule
        \end{tabular*}
    \end{minipage}
\vspace{-4pt}
\end{table*}

\subsection{Portability to Pretrained World Models (Q4)}
\label{sec:sana_train}

\textbf{Settings.} We apply a two-stage adaptation to pretrained SANA-WM~\citep{zhu2026sanawm}, using a subset of its training data~\citep{li2025sekai}.
We fine-tune a bidirectional teacher on data sampled at 10, 20, and 30 FPS for 3k steps, followed by 3k steps of multi-rate Self-Forcing~\citep{huang2025selfforcing} to obtain a streaming student.
We compare the  student with the released SANA-WM Streaming model.
The baseline applies Stride-RoPE at inference with $f_{\mathrm{base}}=16$, whereas the adapted model uses $f_{\mathrm{base}}=30$ for both training and rollout.
Both models are evaluated under the same sampling protocol for SR.
We also report LPIPS as an auxiliary reference for perceptual quality.

\textbf{Results.} As shown in Tab.~\ref{tab:sana_transfer}, LPIPS does not increase at any evaluated rate, providing a complementary quality check after adaptation.
For rate fidelity, except at the base model's native rate of 16 FPS, RAR yields SR values that are closer to 1 or effectively unchanged: it improves the results at 10, 20, and 30 FPS and remains comparable at 24 FPS.
These results suggest that multi-rate adaptation can further improve stride-based rate control in the pretrained model.
The baseline applies Stride-RoPE without finetuning, while rates near its native 16 FPS, such as 20 FPS, require only modest temporal rescaling.
These factors may partly explain why the baseline already achieves reasonable rate fidelity at some non-native rates.
The residual under-scaling at 10 FPS may reflect the difficulty of adapting to larger temporal intervals under a limited training budget, with optical-flow measurement uncertainty potentially contributing to the observed deviations.
Overall, RAR adaptation improves rate fidelity at multiple rollout rates while preserving perceptual quality, supporting its portability.
\section{Conclusion}
We introduce Rate-Adaptive Rollout (RAR) to address temporal-rate coupling in interactive world models.
This coupling encompasses the mismatch between runtime throughput and the target rollout rate, as well as the trade-off between temporal resolution and physical horizon under a fixed budget.
RAR combines physical-time-aligned Stride-RoPE with synchronized multi-stride sampling of videos and controls to learn controlled dynamics across temporal scales and support controllable rollout.
RAR demonstrates consistent generation across training and unseen rates, online rate switching, and an improved quality--horizon trade-off.
These capabilities allow RAR to adapt the rollout rate to runtime throughput while preserving consistent physical-time semantics, offering a practical path toward flexible, real-time interactive world models across hardware conditions.
\newpage
\bibliography{rar}
\bibliographystyle{iclr2027_conference}
\newpage
\appendix
\part*{\mdseries Appendix}

\section{Experimental Details}
\label{app:experimental_details}
We evaluate RAR on Doom, CS2, and FPV. We generate the Doom data using ViZDoom~\citep{Wydmuch2019ViZdoom} and Sample Factory~\citep{petrenko2020sf}, and construct the CS2 data from game demos~\citep{cs2_demo_to_dataset}; both datasets include keyboard-and-mouse controls. For FPV, we use the public Human-Piloted Drone Racing dataset and its camera trajectories~\citep{pfeiffer2021human}.
We initialize the Doom and CS2 models from Matrix-Game 2.0~\citep{he2025matrix}, and initialize the FPV model from SkyReels~\citep{chen2025skyreelsv2infinitelengthfilmgenerative}. After adding action control modules, the Doom and CS2 models contain approximately 1.8B parameters each, while the FPV model contains approximately 1.3B parameters. 
Tab.~\ref{tab:appendix_main_settings} summarizes the main settings for each environment. All training rates within an environment share the same model parameters and fixed sequence length, and therefore correspond to different physical horizons. 
We have also provided the details of the training we conducted in the SANA-WM fine-tuning process (Sec.~\ref{sec:sana_train}).

\begin{center}
\begin{minipage}{\columnwidth}
    \centering
    \captionof{table}{Model and training settings. Bid., DF, and SF denote bidirectional training, Diffusion Forcing, and Self-Forcing, respectively. $F$ is the sequence length in video frames. Bid. and DF entries report steps / learning rate; SF entries report steps / student learning rate / fake-model learning rate.}
    \label{tab:appendix_main_settings}
    \scriptsize
    \resizebox{\linewidth}{!}{%
    \begin{tabular}{lcccccccc}
        \toprule
        Env. & Res. & Params. & $f_{\mathrm{base}}$ & $\mathcal{F}_{\mathrm{train}}$ & $F$ & Bid. & DF & SF \\
        \midrule
        Doom & 360p & 1.8B & 70 & $\{14,35,70\}$ & 213 & $10k / 10^{-4}$ & $12k / 10^{-4}$ & $5k / 10^{-7},\,3\!\times\!10^{-8}$ \\
        CS2 & 360p & 1.8B & 64 & $\{16,32,64\}$ & 241 & $10k / 10^{-4}$ & $10k / 10^{-4}$ & 	$2k / 10^{-7}, 3\times\!10^{-8}$ \\
        FPV & 480p & 1.3B & 60 & $\{20,30,60\}$ & 241 & $15k / 10^{-4}$ & $11k / 10^{-4}$ & $5k / 10^{-6}, 3\!\times\!10^{-7}$ \\
        SANA-WM & 720p & 1.6B & 30 & $\{10,20,30\}$ & 961 & $3k / 10^{-5}$ & --- & 	$3k / 10^{-5}, 10^{-6}$ \\
        \bottomrule
    \end{tabular}}
\end{minipage}
\end{center}

Following Solaris~\citep{solaris2026}, we adopt a similar three-stage training pipeline consisting of bidirectional training, Diffusion Forcing~\citep{chen2024diffusion}, and Self-Forcing~\citep{huang2025selfforcing}. The environment-specific training steps and learning rates are listed in Tab.~\ref{tab:appendix_main_settings}.

\paragraph{Evaluation protocols.}
All evaluations use random seed 42, with matched initial conditions and underlying physical-time control trajectories across methods and rates.
The following summary identifies the evaluated training stages, case counts, and metrics; counts apply per environment.
SF denotes the final Self-Forcing checkpoint, following the training pipeline described above.

\begin{center}
    \small
    \setlength{\tabcolsep}{4pt}
    \renewcommand{\arraystretch}{1.15}
    \begin{tabular}{p{0.34\linewidth}p{0.14\linewidth}p{0.25\linewidth}p{0.17\linewidth}}
        \toprule
        Experiment & Stage & Cases & Metrics \\
        \midrule
        Multi-rate/comparison (Tabs.~\ref{tab:main_results}, \ref{tab:rate_control_comparison}) & SF & 64; SR: 32 & LPIPS, Drift, DOVER, SR \\
        Dynamic switching (Appd.~\ref{app:rate_switch}) & SF & $64\to32\to24$ & Pearson, nRMSE \\
        Rate--horizon comparisons (Tab.~\ref{tab:training_efficiency}; Figs.~\ref{fig:efficiency_fixed_steps_rates}, \ref{fig:efficiency_fixed_steps_length}, \ref{fig:efficiency_fixed_time}) & Bidirectional & 32 & LPIPS \\
        SANA-WM adaptation (Tab.~\ref{tab:sana_transfer}) & SF & 64; SR: 32 & SR, LPIPS \\
        \bottomrule
    \end{tabular}
\end{center}

For the reported SR results, we select the 32 of 64 cases with the lowest GT cross-rate speed coefficient of variation (CV).
Selection uses only GT, not generated outputs, and all methods share the selected cases.
The same physical-time GT trajectory underlies all rates, so its cross-rate speed CV assesses the reliability of optical-flow-based speed measurements.
The filtered switching protocol instead retains the 32 lowest-LPIPS cases, followed by GT reliability filtering to 24 cases; this is distinct from SR selection.
Metric-specific details are provided in Appd.~\ref{app:rate_fidelity} and Appd.~\ref{app:rate_switch}.

\paragraph{Control resampling.}
As clarified in Sec.~\ref{sec:rar}, $\mathbf{a}(t_n^{(s)})$ in Eq.~\ref{eq:multirate_sampling} represents control over the interval between consecutive sampling times.
We construct these controls from the base-rate stream using identical rules at training and interpolated strides:

\begin{center}
    \small
    \setlength{\tabcolsep}{4pt}
    \renewcommand{\arraystretch}{1.15}
    \begin{tabular}{p{0.28\linewidth}p{0.14\linewidth}p{0.50\linewidth}}
        \toprule
        Control type & Environment & Interval representation \\
        \midrule
        Held-key states & Doom, CS2 & Sampled synchronously with the stride \\
        Events (fire/click) & Doom, CS2 & Triggered if pressed at any time within the interval \\
        Mouse increments & Doom, CS2 & Summed over the interval \\
        Camera trajectories & FPV & Reference-relative pose at the sampling time \\
        Non-integer strides & All & Sample positions mapped to the nearest base-rate frame \\
        \bottomrule
    \end{tabular}
\end{center}

Summation preserves cumulative rotation, event aggregation avoids missed triggers, and poses remain aligned with the sampled times; Stride-RoPE supplies the corresponding interval duration.

\section{Rate-Fidelity Evaluation}
\label{app:rate_fidelity}

\subsection{Evaluation Metrics Definition}

Here, we provide the complete definition of Eq.~\ref{eq:rate_fidelity}. Let $m_{k,d}(f,c)$ denote the optical-flow magnitude between frames separated by $k$ steps for video $d\in\{\mathrm{pred},\mathrm{GT}\}$ in case $c$ at temporal rate $f$. Since this interval corresponds to $k/f$ seconds, we define the physical motion speed for that case and rate as
\begin{equation}
    v_d(f,c) = \frac{m_{k,d}(f,c)}{k/f}.
    \label{eq:rate_fidelity_speed}
\end{equation}
To focus on cross-rate scaling rather than the overall motion magnitude of each case, we normalize speeds separately for generated and ground-truth videos.
For each case, the reference speed is the geometric mean over $\mathcal F_{\mathrm{in}}$, the set of evaluated rates within the training range, including both training and interpolation rates:
\begin{equation}
    \bar v_d(c) = \exp\!\left(
    \frac{1}{|\mathcal F_{\mathrm{in}}|}
    \sum_{f'\in\mathcal F_{\mathrm{in}}}\log v_d(f',c)\right).
    \label{eq:rate_fidelity_reference}
\end{equation}
The ratio $v_d(f,c)/\bar v_d(c)$ measures the motion speed at rate $f$ relative to the reference speed of the same case, removing variations in motion magnitude across scenes. We take its median across cases for robust aggregation and divide it by the same statistic on the GT side to reduce systematic biases in optical-flow measurement. This yields the rate fidelity
\begin{equation}
    \mathrm{SR}(f)
    = \frac{\operatorname{median}_{c}\left[v_{\mathrm{pred}}(f,c)/\bar v_{\mathrm{pred}}(c)\right]}
    {\operatorname{median}_{c}\left[v_{\mathrm{GT}}(f,c)/\bar v_{\mathrm{GT}}(c)\right]},
    \label{eq:rate_fidelity_details}
\end{equation}
where a value close to one indicates agreement with the ground-truth cross-rate scaling, rather than absolute speed accuracy. For the main experiments, we estimate optical flow over short generated clips using approximately matched physical-time intervals. At the case level, the main SR results in Tabs.~\ref{tab:main_results} and~\ref{tab:rate_control_comparison} use the 32 of 64 cases with the lowest GT cross-rate speed CV, using only GT and the same cases for all methods. We also remove spatial regions with excessively small or large motion magnitudes, retaining regions with reliable motion signals for rate estimation.

\paragraph{GT-based reliability selection.}
For each case, we compute the coefficient of variation of GT speed across the five in-range rates:
\[
    \mathrm{CV}_{\mathrm{GT}}(c)=
    \frac{\operatorname{std}_{f\in\mathcal F_{\mathrm{in}}}v_{\mathrm{GT}}(f,c)}
    {\operatorname{mean}_{f\in\mathcal F_{\mathrm{in}}}v_{\mathrm{GT}}(f,c)}.
\]
The underlying GT physical motion is shared across rates, so this statistic assesses the cross-rate reliability of optical-flow speed measurements independently of the evaluated model.
The table reports the median CV over each case set, not a pooled statistic over case--rate pairs:

\begin{center}
    \small
    \begin{tabular}{lrrr}
        \toprule
        Case set & Doom & CS2 & FPV \\
        \midrule
        Selected 32 cases & 0.091 & 0.057 & 0.069 \\
        All 64 cases & 0.122 & 0.087 & 0.108 \\
        \bottomrule
    \end{tabular}
\end{center}

The higher CV over the full set motivates the GT-based reliability selection. Below, we also report results over all 64 cases to assess performance without case-level filtering.

\paragraph{Unnormalized speed ratio (RF).}
To measure absolute speed agreement without removing the overall motion scale, we additionally compute
\[
    \mathrm{RF}(f)=\operatorname{median}_{c\in\mathcal C_{64}}
    \left[\frac{v_{\mathrm{pred}}(f,c)}{v_{\mathrm{GT}}(f,c)}\right],
\]
where $\mathcal C_{64}$ contains all 64 evaluation cases for an environment, and predicted and GT speeds are compared over the same physical interval.
We form the ratio within each case before taking the median, without cross-rate normalization or case-level filtering; the ideal value is 1.
Multiplying all predicted speeds by a common positive factor multiplies RF by that factor, whereas it leaves SR unchanged.
RF thus complements SR by exposing global speedup or slowdown rather than only relative cross-rate scaling.

\paragraph{All-case validation.}
We evaluate the final Single, Embed, and RAR models from Appd.~\ref{app:rate_control_comparison} on the same 64 cases per environment.
Alongside RF, we recompute SR using Eq.~\ref{eq:rate_fidelity_details} over all 64 cases, without selecting the GT-reliable subset.
Both metrics have an ideal value of 1; $\dagger$ marks interpolation rates unseen by Embed and RAR, while Single is trained only at each environment's base rate.

\begin{center}
    \small
    \setlength{\tabcolsep}{6pt}
    \begin{tabular}{lrrr|rrr}
        \toprule
        & \multicolumn{3}{c|}{RF, all 64 cases} & \multicolumn{3}{c}{SR, all 64 cases} \\
        FPS & Single & Embed & RAR & Single & Embed & RAR \\
        \midrule
        \multicolumn{7}{c}{\textbf{Doom}} \\
        14 & 0.026 & 1.046 & 0.933 & 0.093 & 0.938 & 0.968 \\
        28$^{\dagger}$ & 0.291 & 1.000 & 0.997 & 0.937 & 0.916 & 0.997 \\
        35 & 0.457 & 1.040 & 1.024 & 1.179 & 0.948 & 1.014 \\
        49$^{\dagger}$ & 0.876 & 1.553 & 0.935 & 2.568 & 1.394 & 0.955 \\
        70 & 1.228 & 0.986 & 1.105 & 3.979 & 0.924 & 1.079 \\
        \midrule
        \multicolumn{7}{c}{\textbf{CS2}} \\
        16 & 0.203 & 1.039 & 0.960 & 0.422 & 0.884 & 0.879 \\
        24$^{\dagger}$ & 0.391 & 1.011 & 0.971 & 0.724 & 0.944 & 0.928 \\
        32 & 0.645 & 1.097 & 1.056 & 1.126 & 0.939 & 0.949 \\
        48$^{\dagger}$ & 0.846 & 1.322 & 1.227 & 1.680 & 1.202 & 1.124 \\
        64 & 0.925 & 1.061 & 1.293 & 1.817 & 0.999 & 1.180 \\
        \midrule
        \multicolumn{7}{c}{\textbf{FPV}} \\
        20 & 0.053 & 0.815 & 0.982 & 0.138 & 1.062 & 1.026 \\
        30 & 0.344 & 0.709 & 0.983 & 0.785 & 0.984 & 0.986 \\
        40$^{\dagger}$ & 0.826 & 0.710 & 1.093 & 1.742 & 0.994 & 1.004 \\
        50$^{\dagger}$ & 0.949 & 0.576 & 1.114 & 2.238 & 0.773 & 1.014 \\
        60 & 1.040 & 0.854 & 1.078 & 2.371 & 1.206 & 0.942 \\
        \bottomrule
    \end{tabular}
\end{center}

We summarize each metric $M\in\{\mathrm{RF},\mathrm{SR}\}$ by $100\,\operatorname{mean}|M-1|$.
Each environment averages its five evaluated rates; overall and interpolation errors average the 15 and six corresponding environment--rate pairs with equal weights, respectively.
Bold indicates the lowest error for each metric and scope.

\begin{center}
    \small
    \setlength{\tabcolsep}{5pt}
    \begin{tabular}{lrrr|rrr}
        \toprule
        & \multicolumn{3}{c|}{RF MAE (\%)} & \multicolumn{3}{c}{All-case SR MAE (\%)} \\
        Scope & Single & Embed & RAR & Single & Embed & RAR \\
        \midrule
        Doom & 51.56 & 13.06 & \textbf{5.27} & 113.91 & 13.35 & \textbf{3.46} \\
        CS2 & 39.79 & \textbf{10.62} & 12.89 & 49.55 & \textbf{8.72} & 10.97 \\
        FPV & 37.34 & 26.74 & \textbf{6.40} & 88.55 & 10.36 & \textbf{2.31} \\
        Overall (15 pairs) & 42.90 & 16.81 & \textbf{8.19} & 84.00 & 10.81 & \textbf{5.58} \\
        Interpolation (6 pairs) & 30.35 & 26.67 & \textbf{8.85} & 76.12 & 16.14 & \textbf{4.36} \\
        \bottomrule
    \end{tabular}
\end{center}

RAR achieves the lowest overall and interpolation RF errors (8.19\% and 8.85\%), supporting absolute speed agreement in addition to relative rate scaling.
Without case-level filtering, RAR also retains its overall and interpolation SR advantages: its errors are 5.58\% and 4.36\%, compared with Embed's 10.81\% and 16.14\%.

\section{Comparison with Alternative Rate-Control Methods}
\label{app:rate_control_comparison}

We compare RAR with two alternative rate-control methods using the same backbone, training window, three-stage training pipeline, optimization-step budgets, and other training settings, except for the training-rate sampling and rate-conditioning mechanisms described below.
All evaluations in this section use the final Self-Forcing checkpoints after all three training stages.
\textbf{Single (Stride-RoPE alone)} is trained only at the base FPS of each environment (70 FPS for Doom, 64 FPS for CS2, and 60 FPS for FPV) and directly applies Stride-RoPE at inference to control the rollout rate.
\textbf{Embed (multi-stride alone)} uses the same multi-rate training data as RAR but retains frame-index-based temporal RoPE and introduces an explicit FPS embedding.
Inspired by FPS conditioning in Stable Video Diffusion~\citep{blattmann2023stable} and VideoCrafter~\citep{chen2023videocrafter1}, the Embed implementation first normalizes the target FPS by the base FPS, encodes the resulting continuous scalar using Fourier embeddings, and adds the FPS embedding to the diffusion timestep embedding.
Single versus RAR tests multi-rate training with Stride-RoPE, while Embed versus RAR compares rate representations on the same multi-rate data.

\begin{table*}[htbp]
    \centering
    \caption{Comparison of rate-control methods using final Self-Forcing models. Single is the Stride-RoPE-alone control; Embed is the multi-stride-alone control with FPS conditioning instead of Stride-RoPE. Entries report $\mathrm{SR}$, whose ideal value is 1. $\dagger$ marks rates unseen by both multi-rate models (Embed and RAR); Single is trained only at the base FPS of each environment. MAE is the equal-weight mean of $|\mathrm{SR}-1|$ over the five displayed rates, expressed as a percentage. Bold indicates the SR closest to 1 in each row or the lowest MAE.}
    \label{tab:rate_control_comparison}
    \footnotesize
    \setlength{\tabcolsep}{2pt}
    \renewcommand{\arraystretch}{1.1}
    \begin{minipage}[t]{0.32\textwidth}
        \centering
        \textbf{(a) Doom}\par\smallskip
        \begin{tabular*}{\linewidth}{@{\extracolsep{\fill}}lrrr@{}}
            \toprule
            FPS & Single & Embed & RAR \\
            \midrule
            14 & 0.106 & 0.921 & \textbf{0.923} \\
            28$^{\dagger}$ & 0.851 & 0.917 & \textbf{1.003} \\
            35 & 1.110 & 0.936 & \textbf{0.999} \\
            49$^{\dagger}$ & 2.483 & 1.395 & \textbf{0.986} \\
            70 & 3.818 & \textbf{0.925} & 1.082 \\
            \midrule
            MAE (\%) & 109.08 & 13.92 & \textbf{3.54} \\
            \bottomrule
        \end{tabular*}
    \end{minipage}\hfill
    \begin{minipage}[t]{0.32\textwidth}
        \centering
        \textbf{(b) CS2}\par\smallskip
        \begin{tabular*}{\linewidth}{@{\extracolsep{\fill}}lrrr@{}}
            \toprule
            FPS & Single & Embed & RAR \\
            \midrule
            16 & 0.412 & 0.900 & \textbf{0.983} \\
            24$^{\dagger}$ & 0.720 & 0.954 & \textbf{1.009} \\
            32 & 1.169 & \textbf{0.974} & 0.970 \\
            48$^{\dagger}$ & 1.678 & 1.181 & \textbf{1.024} \\
            64 & 1.897 & \textbf{1.009} & 1.097 \\
            \midrule
            MAE (\%) & 52.24 & 7.24 & \textbf{3.54} \\
            \bottomrule
        \end{tabular*}
    \end{minipage}\hfill
    \begin{minipage}[t]{0.32\textwidth}
        \centering
        \textbf{(c) FPV}\par\smallskip
        \begin{tabular*}{\linewidth}{@{\extracolsep{\fill}}lrrr@{}}
            \toprule
            FPS & Single & Embed & RAR \\
            \midrule
            20 & 1.151 & 1.051 & \textbf{1.031} \\
            30 & 0.833 & 0.950 & \textbf{1.040} \\
            40$^{\dagger}$ & 1.644 & \textbf{1.009} & 0.960 \\
            50$^{\dagger}$ & 2.134 & 0.844 & \textbf{1.007} \\
            60 & 2.268 & 1.204 & \textbf{0.917} \\
            \midrule
            MAE (\%) & 67.28 & 9.40 & \textbf{4.02} \\
            \bottomrule
        \end{tabular*}
    \end{minipage}
\end{table*}

Tab.~\ref{tab:rate_control_comparison} shows that RAR enables more consistent multi-rate generation than Single, highlighting the benefit of multi-rate training over inference-only temporal rescaling.
RAR also achieves lower overall SR MAE than Embed in all three environments, with a larger aggregate advantage at unseen rates.
Averaged over the nine seen and six unseen environment--rate pairs, Embed and RAR obtain seen-rate MAEs of 7.31\% and 5.09\%, respectively, but unseen-rate MAEs of 14.50\% and 1.62\%.
This gap indicates that Embed does not interpolate as reliably to unseen FPS values despite its continuous FPS embedding, whereas RAR better preserves cross-rate temporal consistency.

\paragraph{Generation quality after causal conversion.}
We additionally compare the final Single and RAR models using LPIPS, with the same models and evaluation settings as in Tab.~\ref{tab:rate_control_comparison}.
The comparison uses all 64 cases per environment and seed 42, with the matched training windows and update budgets described above.
The table reports per-rate LPIPS (lower is better) and its equal-weight mean over the five rates; the RAR entries are those in Tab.~\ref{tab:main_results}.
Here, $\dagger$ marks rates unseen by RAR, and bold indicates the lower LPIPS in each pair.

\begin{center}
    \small
    \setlength{\tabcolsep}{3pt}
    \renewcommand{\arraystretch}{1.1}
    \begin{minipage}[t]{0.32\linewidth}
        \centering
        \textbf{(a) Doom}\par\smallskip
        \begin{tabular*}{\linewidth}{@{\extracolsep{\fill}}lrr@{}}
            \toprule
            FPS & Single & RAR \\
            \midrule
            14 & 0.4468 & \textbf{0.4279} \\
            28$^{\dagger}$ & 0.4291 & \textbf{0.4179} \\
            35 & 0.4180 & \textbf{0.3991} \\
            49$^{\dagger}$ & 0.4124 & \textbf{0.4003} \\
            70 & 0.3988 & \textbf{0.3852} \\
            \midrule
            Mean & 0.4210 & \textbf{0.4061} \\
            \bottomrule
        \end{tabular*}
    \end{minipage}\hfill
    \begin{minipage}[t]{0.32\linewidth}
        \centering
        \textbf{(b) CS2}\par\smallskip
        \begin{tabular*}{\linewidth}{@{\extracolsep{\fill}}lrr@{}}
            \toprule
            FPS & Single & RAR \\
            \midrule
            16 & 0.6815 & \textbf{0.6489} \\
            24$^{\dagger}$ & 0.6566 & \textbf{0.6200} \\
            32 & 0.6526 & \textbf{0.5980} \\
            48$^{\dagger}$ & 0.6305 & \textbf{0.5944} \\
            64 & 0.6055 & \textbf{0.5823} \\
            \midrule
            Mean & 0.6453 & \textbf{0.6087} \\
            \bottomrule
        \end{tabular*}
    \end{minipage}\hfill
    \begin{minipage}[t]{0.32\linewidth}
        \centering
        \textbf{(c) FPV}\par\smallskip
        \begin{tabular*}{\linewidth}{@{\extracolsep{\fill}}lrr@{}}
            \toprule
            FPS & Single & RAR \\
            \midrule
            20 & 0.6544 & \textbf{0.5247} \\
            30 & 0.5962 & \textbf{0.5161} \\
            40$^{\dagger}$ & 0.5465 & \textbf{0.5011} \\
            50$^{\dagger}$ & 0.5150 & \textbf{0.5035} \\
            60 & 0.5128 & \textbf{0.4944} \\
            \midrule
            Mean & 0.5650 & \textbf{0.5080} \\
            \bottomrule
        \end{tabular*}
    \end{minipage}
\end{center}

RAR achieves lower LPIPS than Single at all 15 environment--rate pairs, including Single's native rates of 70, 64, and 60 FPS on Doom, CS2, and FPV, respectively.
Thus, under matched training-update budgets, the lower-rate quality gains and preservation of high-rate quality survive Diffusion Forcing and Self-Forcing conversion, consistent with the bidirectional findings in Tab.~\ref{tab:training_efficiency} (top).

\section{Extrapolation Beyond the Training Range}
\label{app:extrapolation}
Tab.~\ref{tab:extrapolation} reports the rate fidelity of rollout rates outside the training range. For comparison, the table also lists the interpolation rates and the intermediate training rate of each environment. All SR values at extrapolated rates are computed per case adapting Eq.~\ref{eq:rate_fidelity_details} for extrapolation: the GT normalization denominator is the geometric mean of the in-scope GT speeds for the same case, and the extrapolated rate is not included in this reference. The per-case ratios are formed first and then aggregated by their median.
At the low-rate side, the rollout stride exceeds the largest stride observed during training: 7 FPS on Doom, 8 FPS on CS2, and 10 FPS on FPV correspond to strides of 10, 8, and 6, respectively. Their SR values are 0.739, 0.636, and 0.634, all clearly below one, indicating under-scaling of motion relative to the ground-truth cross-rate trend at these rates.
At the high-rate side, Doom at 84 FPS and CS2 at 72 FPS yield SR values of 1.191 and 1.273, respectively, indicating overshoot, whereas FPV at 80 FPS yields 0.864 and therefore undershoots. The direction of the out-of-range bias thus depends on the environment and extrapolation regime rather than following a universal pattern. These results lie outside the scope of our claims and are reported only to characterize the boundary of the method.

\noindent
\begin{minipage}{\columnwidth}
\captionof{table}{Rate fidelity at out-of-range rollout rates. Out-of-range rates are listed together with in-range interpolation and training rates for comparison. Out-of-range rates lie outside the scope of our claims.}
\label{tab:extrapolation}
\vspace{0.4\baselineskip}
\centering
\setlength{\tabcolsep}{6pt}
\begin{tabular*}{0.6\columnwidth}{@{\extracolsep{\fill}}p{0.2\columnwidth}p{0.06\columnwidth}p{0.08\columnwidth}p{0.1\columnwidth}@{}}
    \toprule
    Environment & FPS & Type & $\mathrm{SR}\!\rightarrow\!1$ \\
    \midrule
    Doom & 7  & \textit{Extrap.} & 0.739 \\
         & 28 & \textbf{Interp.} & 1.003 \\
         & 35 & Train   & 0.999 \\
         & 49 & \textbf{Interp.} & 0.986 \\
         & 84 & \textit{Extrap.} & 1.191 \\
    \midrule
    CS2  & 8  & \textit{Extrap.} & 0.636 \\
         & 24 & \textbf{Interp.} & 1.009 \\
         & 32 & Train   & 0.970 \\
         & 48 & \textbf{Interp.} & 1.024 \\
         & 72 & \textit{Extrap.} & 1.273 \\
    \midrule
    FPV  & 10 & \textit{Extrap.} & 0.634 \\
         & 20 & Train   & 1.031 \\
         & 40 & \textbf{Interp.} & 0.960 \\
         & 50 & \textbf{Interp.} & 1.007 \\
         & 80 & \textit{Extrap.} & 0.864 \\
    \bottomrule
\end{tabular*}
\end{minipage}

\section{Dynamic Rollout-Rate Switching}
\label{app:rate_switch}
\begin{figure*}
    \centering
    \includegraphics[width=0.6\textwidth]{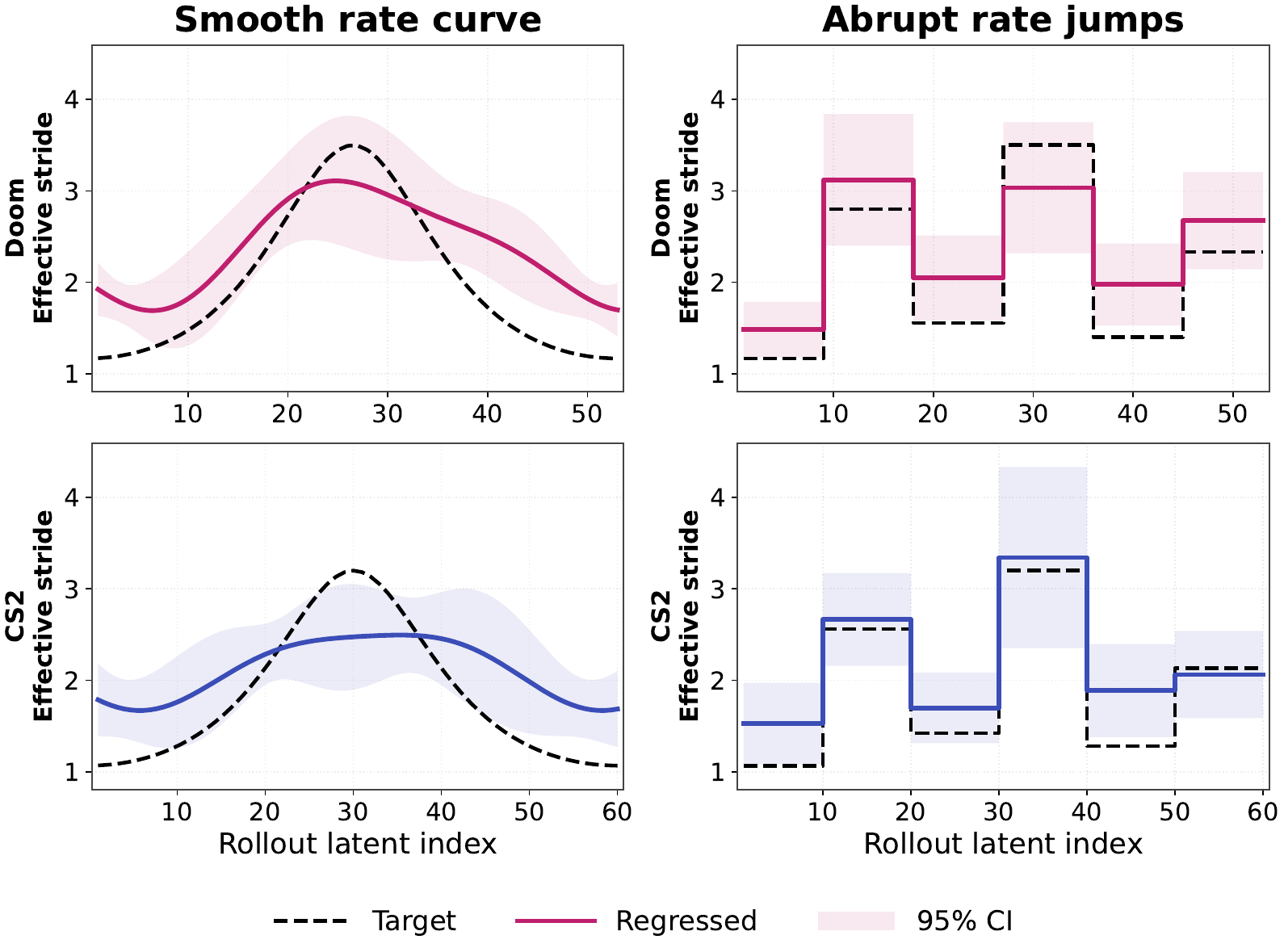}
    \caption{Additional dynamic rollout-rate switching results on Doom and CS2. The evaluation protocol and plot conventions follow Fig.~\ref{fig:rate_switch_main}.}
    \label{fig:rate_switch_appendix}
\end{figure*}

We evaluate the final models on 64 rollouts per environment and schedule, using seed 42 and the smooth and abrupt schedules described in the main text.
We first describe the processed protocol used in Figs.~\ref{fig:rate_switch_main} and~\ref{fig:rate_switch_appendix}, then provide all-case and uncalibrated validation.

\paragraph{Processed schedule-tracking evaluation.}
Because absolute motion scales vary across scenes, we apply a single global multiplicative calibration to each rollout, aligning the geometric mean of the measured motion with that of the target stride. This calibration only matches the vertical scale and preserves the relative temporal variation of the curve. The evaluation therefore measures how well the model tracks a dynamic schedule rather than its absolute stride calibration. For the smooth schedule, we first group adjacent intervals into bins and then fit each rollout using a second-order periodic harmonic regression. For the abrupt schedule, we perform piecewise-constant regression over its six constant-stride segments.

To reduce interference from low-quality generations and unreliable motion measurements, we apply a two-stage case filtering procedure. First, for each environment and schedule, we retain the 32 of 64 cases with the lowest mean LPIPS between the GT and generated rollouts. We then assess measurement reliability using the tracking nRMSE obtained from the GT videos under the same protocol and retain 24 cases with lower measurement errors. These criteria evaluate generation quality and measurement reliability, respectively, without selecting cases according to the generated rollouts' final tracking scores. The solid curves and shaded regions show the mean regressed response over the retained cases and its 95\% confidence interval, respectively.

As shown in Fig.~\ref{fig:rate_switch_appendix}, both Doom and CS2 recover the overall variation of the target stride under the smooth and abrupt schedules. On Doom, the smooth and abrupt schedules achieve Pearson correlations of 0.941 and 0.938, with nRMSE values of 0.219 and 0.185, respectively. The corresponding results on CS2 are 0.899 and 0.963 for Pearson correlation and 0.248 and 0.157 for nRMSE. In particular, under the abrupt schedule, both environments correctly recover the alternating increases and decreases across all six segments, showing that RAR responds to discrete rollout-rate switches without interrupting the rollout.

Compared with FPV in the main paper, the smooth responses on Doom and CS2 are flatter and have wider confidence intervals. One possible reason is that observable motion in these game environments primarily arises from intermittent turns, firing events, or short movements, rather than the persistent motion provided by the continuous camera trajectories in FPV. Consequently, local optical flow reflects both the target stride and content-dependent motion, making gradual changes in the smooth schedule harder to isolate. The GT-side diagnostic supports this interpretation: under the same measurement protocol, the GT Pearson correlations for the smooth schedule are only 0.841 on Doom and 0.896 on CS2, indicating that the optical-flow measurement cannot fully recover the target curve even without generation errors. In contrast, the abrupt schedule aggregates motion over longer constant-stride segments, increasing the GT Pearson correlations to 0.931 and 0.996, respectively, while also yielding lower nRMSE for the generated rollouts. The residual discrepancies in the figure should therefore be attributed partly to measurement uncertainty caused by intermittent motion, rather than being interpreted entirely as errors in rollout-rate control.

\paragraph{All-case and uncalibrated validation.}
We assess whether schedule tracking persists without the above processing steps.
The first three result columns below retain the original displacement estimator and multiplicative calibration: the first uses the processed protocol, the second includes all 64 rollouts, and the third additionally removes regression.
All three use mean response curves across their respective case sets.

The final column is a separate GT-referenced estimate of effective stride, not merely another ablation of the same estimator.
For each latent interval, we divide generated displacement by the GT displacement per base-frame interval over the same physical interval; equivalently,
\[
    \widehat{s}(c,k)=s_k\frac{D_{\mathrm{pred}}(c,k)}{D_{\mathrm{GT}}(c,k)},
\]
where $s_k$ is the target stride and $D_{\mathrm{pred}}(c,k)$ and $D_{\mathrm{GT}}(c,k)$ are the optical-flow displacements for case $c$ over the same interval $k$.
We take the median across all 64 rollouts, with no case filtering, regression, or fitted scale factor.
The abrupt-schedule scores summarize the six constant-stride segments.
The first three columns therefore use mean curves, whereas the final column uses a median curve with a different stride estimator.
Cells report Pearson correlation with the target schedule and nRMSE, respectively.

\begin{center}
    \footnotesize
    \setlength{\tabcolsep}{4pt}
    \renewcommand{\arraystretch}{1.15}
    \begin{tabular}{llcccc}
        \toprule
        Env. & Schedule & \shortstack{Processed\\($n=24$)} & \shortstack{No filtering\\($n=64$)} & \shortstack{+ No regression\\($n=64$)} & \shortstack{GT-referenced,\\uncalibrated ($n=64$)} \\
        \midrule
        Doom & Smooth & 0.941 / 0.219 & 0.967 / 0.282 & 0.927 / 0.291 & 0.867 / 0.175 \\
        Doom & Abrupt & 0.938 / 0.185 & 0.976 / 0.211 & 0.977 / 0.213 & 0.982 / 0.074 \\
        CS2 & Smooth & 0.899 / 0.248 & 0.945 / 0.328 & 0.937 / 0.353 & 0.856 / 0.192 \\
        CS2 & Abrupt & 0.963 / 0.157 & 0.968 / 0.277 & 0.970 / 0.281 & 0.860 / 0.197 \\
        FPV & Smooth & 0.998 / 0.169 & 0.994 / 0.277 & 0.988 / 0.307 & 0.997 / 0.108 \\
        FPV & Abrupt & 0.970 / 0.122 & 0.973 / 0.231 & 0.974 / 0.229 & 0.980 / 0.118 \\
        \bottomrule
    \end{tabular}
\end{center}

Removing case filtering increases Pearson correlation in five of six settings, while nRMSE rises to 0.211--0.328, consistent with the exclusion of poorer generations and less reliable measurements in the processed protocol.
After regression is also removed, Pearson remains 0.927--0.988.
The GT-referenced, uncalibrated estimates yield Pearson 0.856--0.997 and nRMSE 0.074--0.197: the former demonstrates schedule tracking, while the latter quantifies amplitude deviations from the target.
Thus, RAR tracks the schedule without case filtering, regression, or scale fitting.

\begin{figure*}[htbp]
    \centering
    \includegraphics[width=0.96\textwidth]{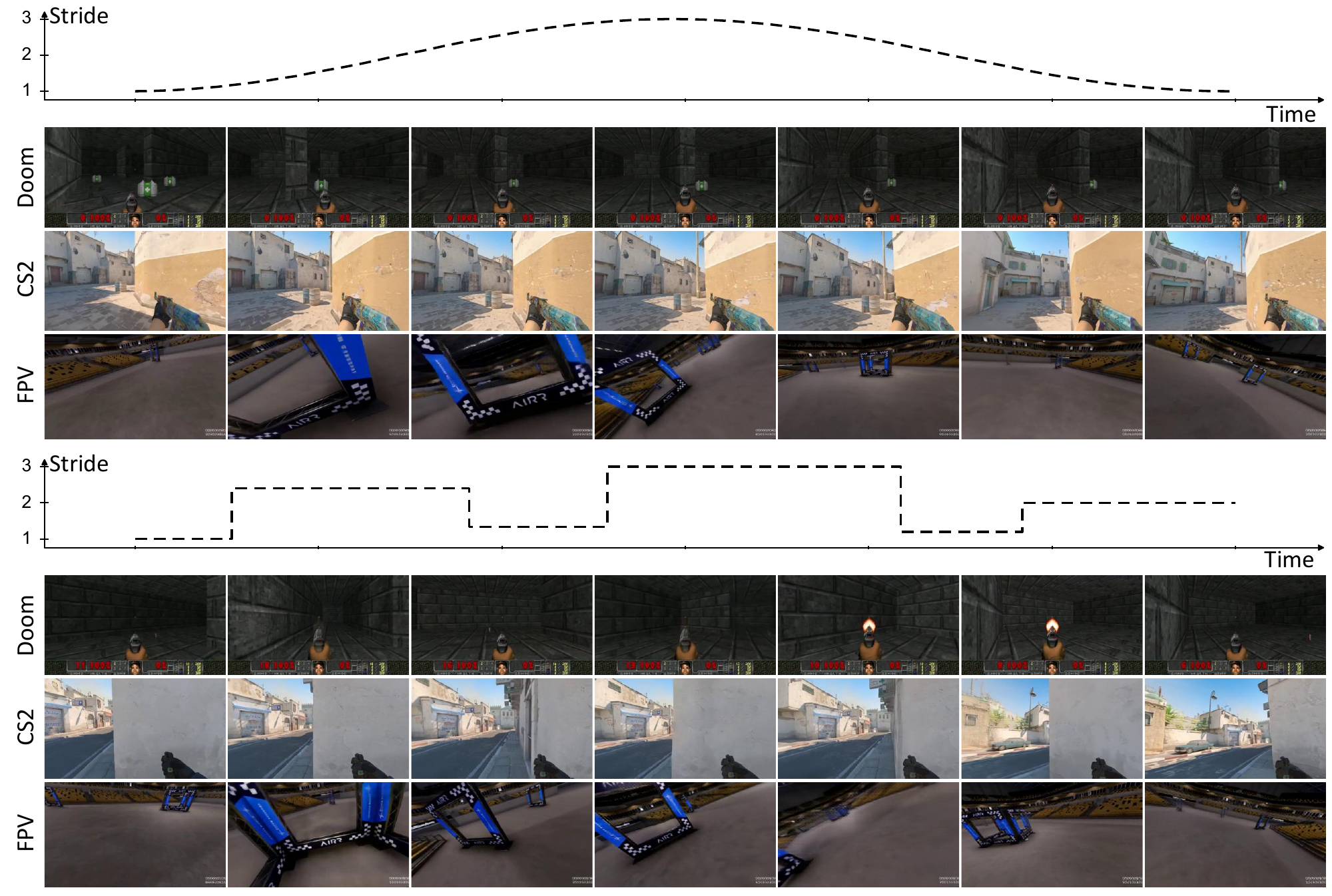}
    \caption{Qualitative dynamic rollout-rate switching on Doom, CS2, and FPV. The upper and lower groups show generated frames under smooth and abrupt schedules, respectively.}
    \label{fig:rate_switch_qualitative}
\end{figure*}

\paragraph{Qualitative results.}
The qualitative examples in Fig.~\ref{fig:rate_switch_qualitative} show no noticeable degradation in visual quality during smooth or abrupt rollout-rate changes on Doom, CS2, and FPV.

\section{Rate--Horizon Trade-off: Additional Results}
\label{app:rate_horizon}

\subsection{Bidirectional Training Analysis}
\label{app:bidirectional_rate_horizon}
The following analyses expand the bidirectional comparisons in Tab.~\ref{tab:training_efficiency} on Doom; final causal models are evaluated separately in Appd.~\ref{app:causal_long_rollout}.

\begin{figure*}
    \centering
    \includegraphics[width=0.96\textwidth]{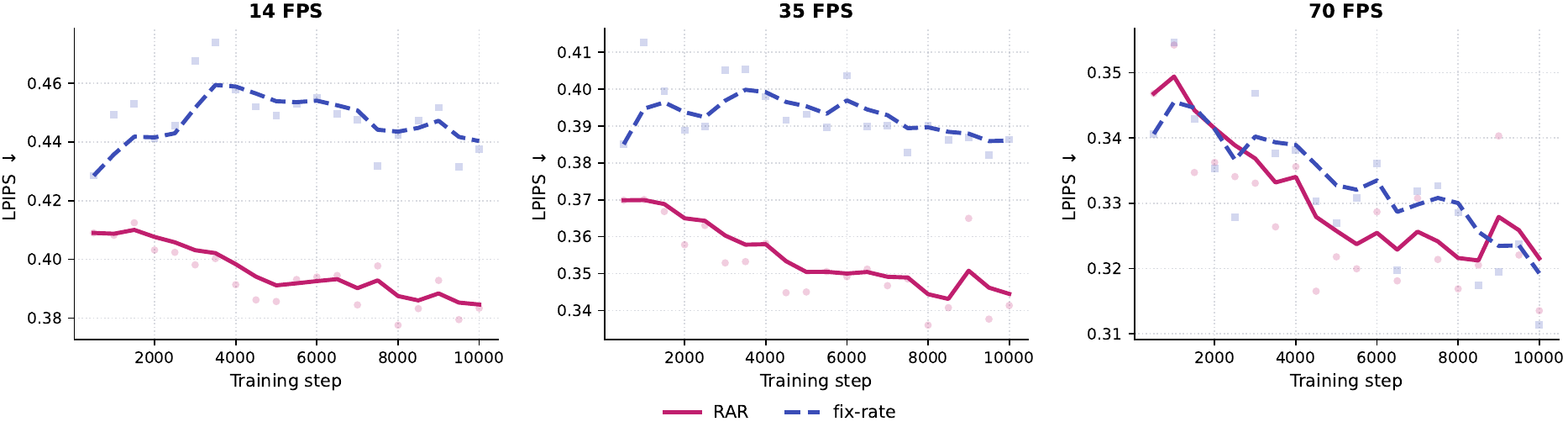}
    \caption{LPIPS under a fixed training-update budget on Doom. RAR is jointly trained at 14, 35, and 70 FPS, whereas the fixed-rate baseline is trained only at 70 FPS. Light markers denote individual checkpoints, and curves show the corresponding trends.}
    \label{fig:efficiency_fixed_steps_rates}
\end{figure*}

\noindent\textbf{Fixed training steps.}
We first compare RAR with the fixed-rate baseline under the same number of training updates. Both start from the same bidirectional initialization and use the same model architecture and 213-frame training window; RAR is jointly trained with data at 14, 35, and 70 FPS, whereas the fixed-rate baseline uses only 70-FPS data. Fig.~\ref{fig:efficiency_fixed_steps_rates} plots LPIPS over training at the three evaluation rates. The curves exhibit a clear training trend. At 14 and 35 FPS, the LPIPS of RAR decreases steadily, while the fixed-rate baseline remains at a substantially higher level after its initial fluctuations and shows no corresponding learning trend at the lower rates. At 70 FPS, both curves decrease similarly and gradually converge later in training. This indicates that RAR progressively acquires lower-rate generation capabilities during training, rather than benefiting from fluctuations at a particular checkpoint, while preserving its generation capability at the highest training rate. For reference, after 10k updates, the LPIPS values of RAR at 14 and 35 FPS are 0.3834 and 0.3413, compared with 0.4376 and 0.3863 for the fixed-rate baseline. At 70 FPS, the two models obtain comparable values of 0.3135 and 0.3113, respectively.

\begin{figure*}
    \centering
    \includegraphics[width=0.7\textwidth]{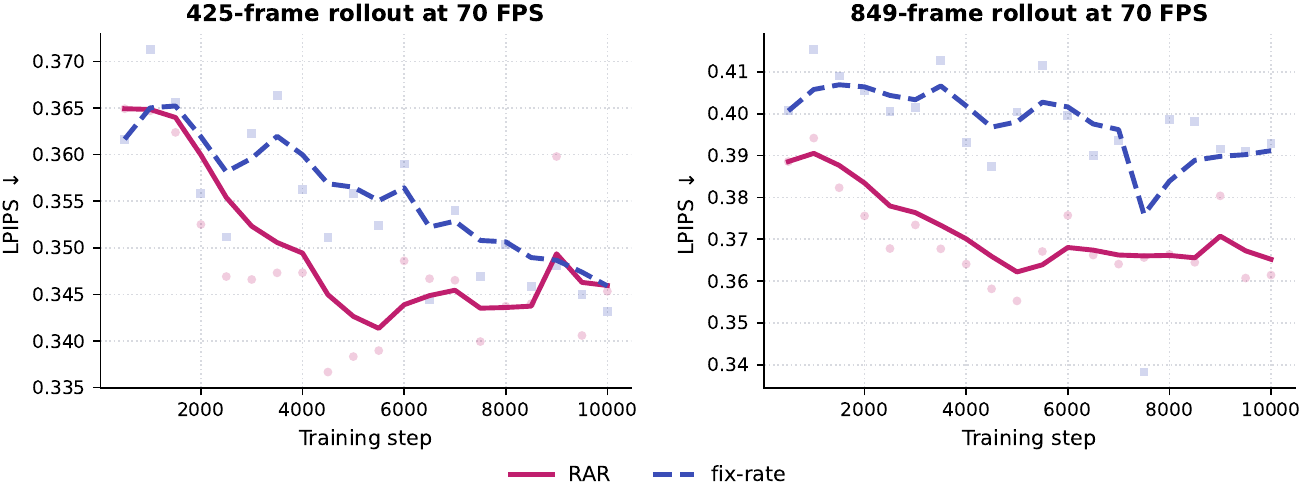}
    \caption{Long-rollout LPIPS under a fixed training-update budget on Doom. Both models are trained with 213-frame windows and evaluated at 70 FPS using 425- and 849-frame rollouts.}
    \label{fig:efficiency_fixed_steps_length}
\end{figure*}

We further evaluate both 213-frame-trained models on longer rollouts by jointly denoising the full sequence, without changing their training windows.
Fig.~\ref{fig:efficiency_fixed_steps_length} compares their LPIPS when generating 425 and 849 frames at 70 FPS. For 425-frame rollouts, RAR improves more rapidly during the first half of training and reaches lower LPIPS earlier; with further training, the two curves become close, reaching 0.3383 for RAR and 0.3432 for the fixed-rate baseline after 10k updates. On the more challenging 849-frame rollouts, RAR achieves lower LPIPS from the early stages of training and maintains a consistent advantage thereafter, reaching 0.3552 after 10k updates compared with 0.3929 for the fixed-rate baseline. The transition from faster convergence at 425 frames to a persistent gap at 849 frames shows that the benefit of multi-rate training becomes more pronounced as the evaluation rollout length increases. These results indicate that the different physical horizons induced by the training rates improve long-horizon generation stability, allowing RAR to extend to rollouts substantially longer than its training window without increasing the training sequence length.

\noindent\textbf{Matched training time.}
Finally, under approximately matched wall-clock training time at the bidirectional stage, we compare RAR trained with a 213-frame window against fixed-rate models trained directly with 425- and 849-frame windows. Each model is evaluated at 70 FPS using the corresponding long-rollout length.
All three configurations use eight H20 GPUs with batch size 1 per GPU. The table lists the bidirectional checkpoints used in Tab.~\ref{tab:training_efficiency} (bottom), with approximately matched cumulative training times:

\begin{center}
    \small
    \setlength{\tabcolsep}{5pt}
    \renewcommand{\arraystretch}{1.1}
    \begin{tabular}{lrrrr}
        \toprule
        Configuration & \shortstack{Training window\\(frames)} & s/update & \shortstack{Checkpoint\\(updates)} & \shortstack{Approx. training\\time (h)} \\
        \midrule
        RAR & 213 & 80 & 10k & 222 \\
        Fixed rate & 425 & 125 & 6.6k & 229 \\
        Fixed rate & 849 & 420 & 2k & 233 \\
        \bottomrule
    \end{tabular}
\end{center}

\begin{figure*}
    \centering
    \includegraphics[width=0.7\textwidth]{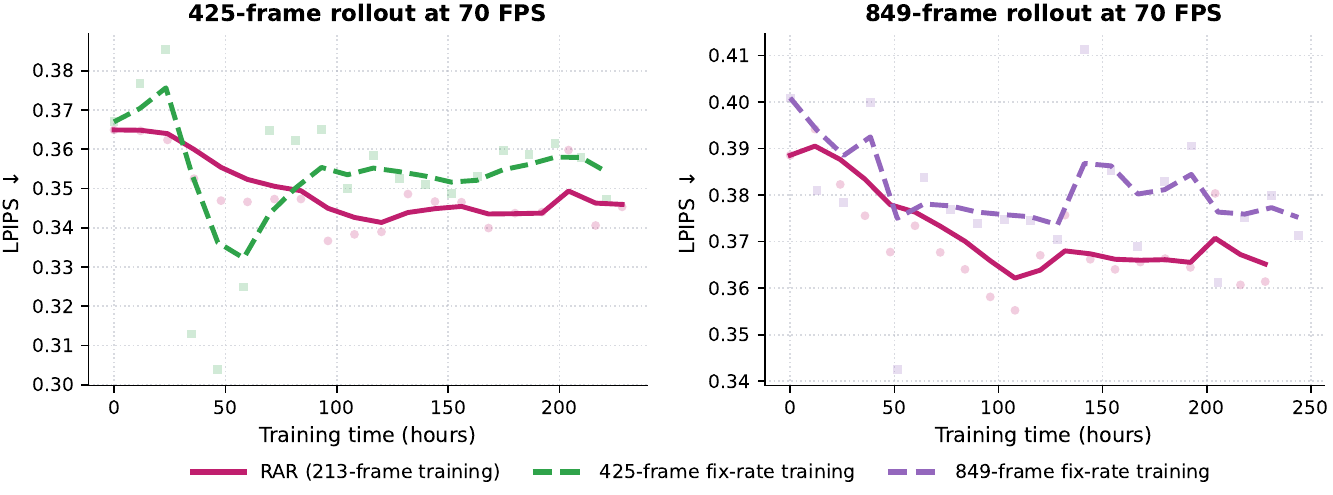}
    \caption{LPIPS of bidirectional models under approximately matched wall-clock training time on Doom. RAR is trained with 213-frame multi-rate sequences and compared with fixed-rate models trained directly using 425- or 849-frame sequences at 70 FPS. Each model is evaluated at its indicated rollout length.}
    \label{fig:efficiency_fixed_time}
\end{figure*}

Total training times differ by no more than 5\%. Each paired comparison uses the same 32 evaluation cases and seed 42, generating either 425 or 849 frames at 70 FPS.
As shown in Fig.~\ref{fig:efficiency_fixed_time}, on the 425-frame evaluation, the LPIPS of RAR decreases steadily with training time and reaches a level comparable to or lower than that of 425-frame fixed-rate training during the middle and later stages, whereas the latter exhibits larger fluctuations without establishing a persistent advantage. On the 849-frame evaluation, RAR reaches lower LPIPS earlier and maintains a clear gap from the 849-frame fixed-rate model over most of the training period. This trend shows that the efficiency advantage of RAR persists throughout training under the same wall-clock budget, rather than appearing only at a particular final checkpoint. Compared with simply increasing the training sequence length, RAR covers multiple physical horizons by jointly training on different rates using shorter sequences and a lower cost per update, yielding comparable or better long-horizon generation quality within the same training time.

\subsection{Long Rollouts of Final Causal Models}
\label{app:causal_long_rollout}
We next test whether the quality--horizon benefit persists after causal conversion and accumulated autoregressive errors.
We use the final Single and RAR models from Appd.~\ref{app:rate_control_comparison}, with matched backbones, training windows, and update counts, and all three training stages completed.
Within each environment, both models use identical KV-cache windows and sink settings, with matched initial conditions, control trajectories, and seed 42.
They autoregressively generate 425 and 849 frames at the base rate of each environment: 70 FPS on Doom, 64 FPS on CS2, and 60 FPS on FPV.
Unlike the bidirectional evaluations above, these rollouts use the final Self-Forcing checkpoints and do not jointly denoise the entire sequence.
The table reports LPIPS, with lower values in bold:

\begin{center}
    \small
    \setlength{\tabcolsep}{5pt}
    \renewcommand{\arraystretch}{1.1}
    \begin{tabular}{lrrrr}
        \toprule
        & \multicolumn{2}{c}{425 frames} & \multicolumn{2}{c}{849 frames} \\
        Environment & Single & RAR & Single & RAR \\
        \midrule
        Doom & 0.4143 & \textbf{0.4087} & 0.4311 & \textbf{0.4306} \\
        CS2 & 0.684 & \textbf{0.529} & 0.808 & \textbf{0.640} \\
        FPV & 0.5238 & \textbf{0.4697} & 0.5720 & \textbf{0.5387} \\
        \bottomrule
    \end{tabular}
\end{center}

RAR achieves lower LPIPS at both rollout lengths in all three environments, with larger gaps on CS2 and FPV.
Together with the per-rate quality comparisons in Appd.~\ref{app:rate_control_comparison}, these results show that the quality--horizon advantage survives Diffusion Forcing and Self-Forcing conversion as well as autoregressive error accumulation under matched training-update budgets.
This extends the quality benefit to final causal models, while the matched-training-time efficiency evidence remains the bidirectional comparison in Appd.~\ref{app:bidirectional_rate_horizon}.

\end{document}